\documentclass[runningheads]{llncs}
\usepackage{graphicx}
\usepackage{algorithm}
\usepackage{multirow}
\usepackage{subfigure}
\usepackage{booktabs}
\usepackage{tikz}
\usepackage{comment}
\usepackage{amsmath,amssymb} % define this before the line numbering.
\usepackage{color}
\usepackage[normalem]{ulem}

\usepackage[accsupp]{axessibility}  % Improves PDF readability for those with disabilities.
\usepackage[pagebackref,breaklinks,colorlinks]{hyperref}

\newcommand{\ie}{\textit{i}.\textit{e}.}
\newcommand{\eg}{\textit{e}.\textit{g}.}
\newcommand{\otsp}{OTSP}

\begin{document}
\pagestyle{headings}
\mainmatter
\def\ECCVSubNumber{986}  % Insert your submission number here

%\title{Object Detection via Single Point Supervision}
\title{Point-to-Box Network for Accurate Object Detection via Single Point Supervision}

\begin{comment}
\titlerunning{ECCV-22 submission ID \ECCVSubNumber} 
\authorrunning{ECCV-22 submission ID \ECCVSubNumber} 
\author{Anonymous ECCV submission}
\institute{Paper ID \ECCVSubNumber}
\end{comment}
% %******************

% \begin{comment}
\titlerunning{Point-to-Box Network for PSOD}
\author{Pengfei Chen\inst{1} \and
Xuehui Yu\inst{1}\and
Xumeng Han\inst{1} \and
Najmul Hassan\inst{2} \and
Kai Wang\inst{2} \and
Jiachen Li\inst{2} \and
Jian Zhao\inst{3} \and
Humphrey Shi\inst{2,4} \and
Zhenjun Han\inst{1}\thanks{Corresponding author.} \and
Qixiang Ye\inst{1}}

\authorrunning{Pengfei Chen et al.}
\institute{University of Chinese Academy of Sciences, Beijing, China \and SHI Lab @ U of Oregon \& UIUC, USA \and  
Institute of North Electronic Equipment, Beijing, China \and
Picsart AI Research (PAIR) \\
\email{\{chenpengfei20, yuxuehui17, hanxumeng19\}@mails.ucas.ac.cn}  \email{\{najmulhassan1628, kk94wang, chrisleesjtu, shihonghui3\}@gmail.com}\\
\email{zhaojian90@u.nus.edu, \email{\{hanzhj, qxye\}@ucas.ac.cn}}}
% \end{comment}
%******************
%******************

\maketitle
%\vspace{-0.2cm}
\begin{abstract}
    Object detection using single point supervision has received increasing attention over the years. However, the performance gap between point supervised object detection (PSOD) and bounding box supervised detection remains large.
    In this paper, we attribute such a large performance gap to the failure of generating high-quality proposal bags which are crucial for multiple instance learning (MIL).
    To address this problem, we introduce a lightweight alternative to the off-the-shelf proposal (OTSP) method and thereby create the Point-to-Box Network (P2BNet), which can construct an inter-objects balanced proposal bag by generating proposals in an anchor-like way. By fully investigating the accurate position information, P2BNet further constructs an instance-level bag, avoiding the mixture of multiple objects. Finally, a coarse-to-fine policy in a cascade fashion is utilized to improve the IoU between proposals and ground-truth (GT). Benefiting from these strategies, P2BNet is able to produce high-quality instance-level bags for object detection. 
    P2BNet improves the mean average precision (AP) by more than 50\% relative to the previous best PSOD method on the MS COCO dataset.
    It also demonstrates the great potential to bridge the performance gap between point supervised and bounding-box supervised detectors. The code will be released at \textcolor{magenta}{\url{github.com/ucas-vg/P2BNet}}.
\end{abstract}

%---------------------------------------------------------------------------------
% %\vspace{-0.2cm}
\section{Introduction}
%\vspace{-0.2cm}
Object detectors ~\cite{fastrcnn,FasterRCNN,DBLP:yolo,DBLP:SSD,DBLP:retinanet_focalloss,DBLP:DETR,DBLP:sparsercnn,scalematch} trained with accurate bounding box annotations have been well received in academia and industry. However, collecting quality bounding box annotations requires extensive human efforts. To solve this problem, weakly supervised object detection~\cite{DBLP:wsddn,DBLP:oicr,DBLP:pcl,DBLP:slv,DBLP:MELM,DBLP:cam,DBLP:wccn,DBLP:Acol} (WSOD) replace bounding box annotations using low-cost image-level annotations. However, lacking crucial location information and experiencing the difficulty of distinguishing dense objects, WSOD methods perform poorly in complex scenarios. Point supervised object detection (PSOD), on the other hand, can provide distinctive location information about the object and is much cheaper compared with that via bounding box supervision.

\begin{figure}[t]
    \centering
    \includegraphics[width=1.\linewidth]{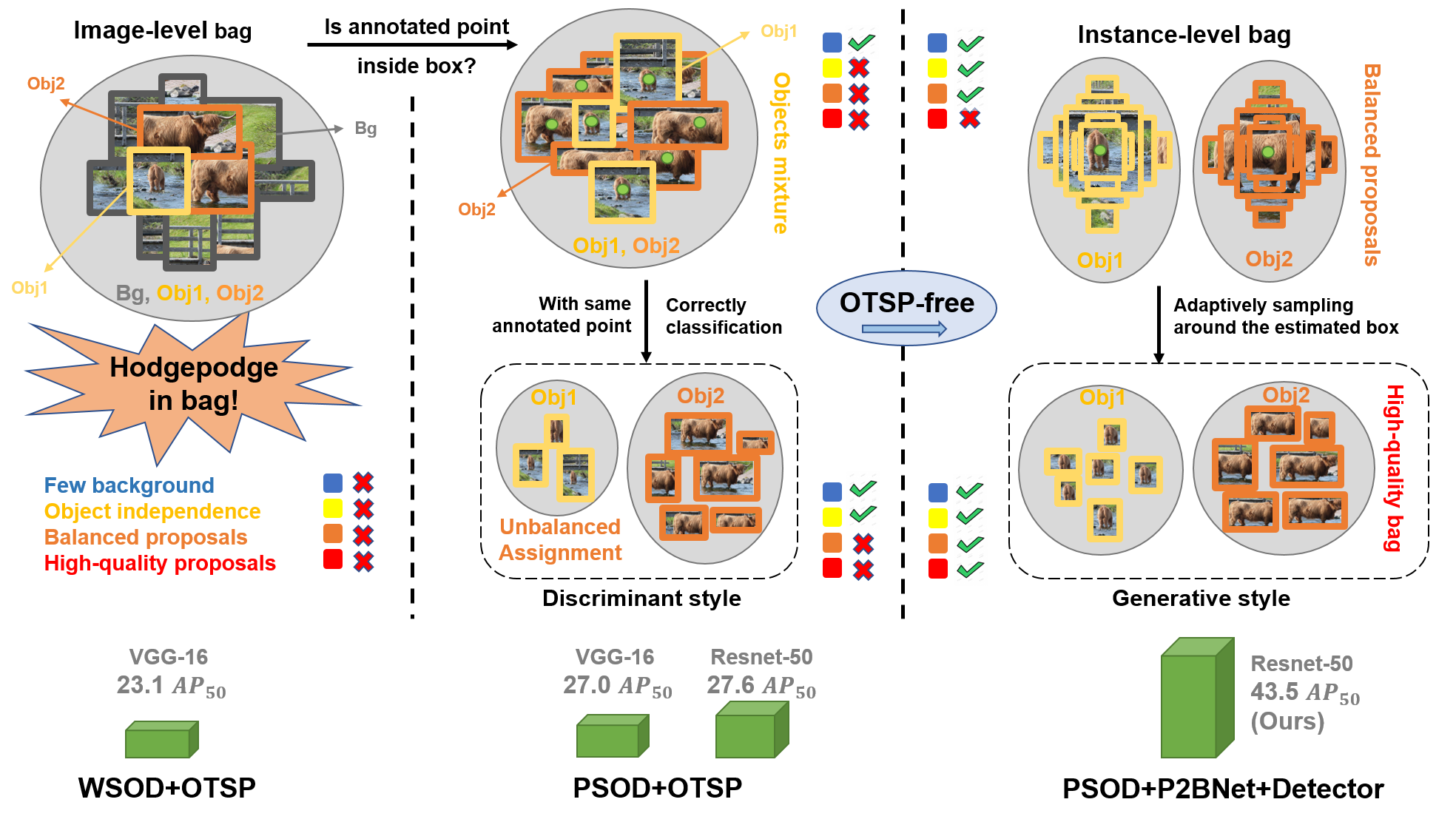}
    %\vspace{-0.7cm}
    \caption{Based on \otsp~methods, the image-level bag in WSOD shows many problems: Too much background, mixture of different objects, unbalanced and low-quality proposals. With point annotation, the previous work UFO$^2$ filters most background in first stage and splits bags for different objects in refinement. Our P2BNet produces balanced instance-level bags in coarse stage and improves bag quality improves by adaptively sampling proposal boxes around the estimated box of the former stage for better optimization. The performance is the performance in COCO-14. The 27.6 AP$_{50}$ is conducted on UFO$^2$ with ResNet-50 and our point annotation for a fair comparison.}
    \label{fig:motivation}
%\vspace{-0.5cm}
\end{figure}

\begin{figure}[t]
%\vspace{-0.2mm}
\centering
\subfigure[]{
\includegraphics[height=.22\columnwidth]{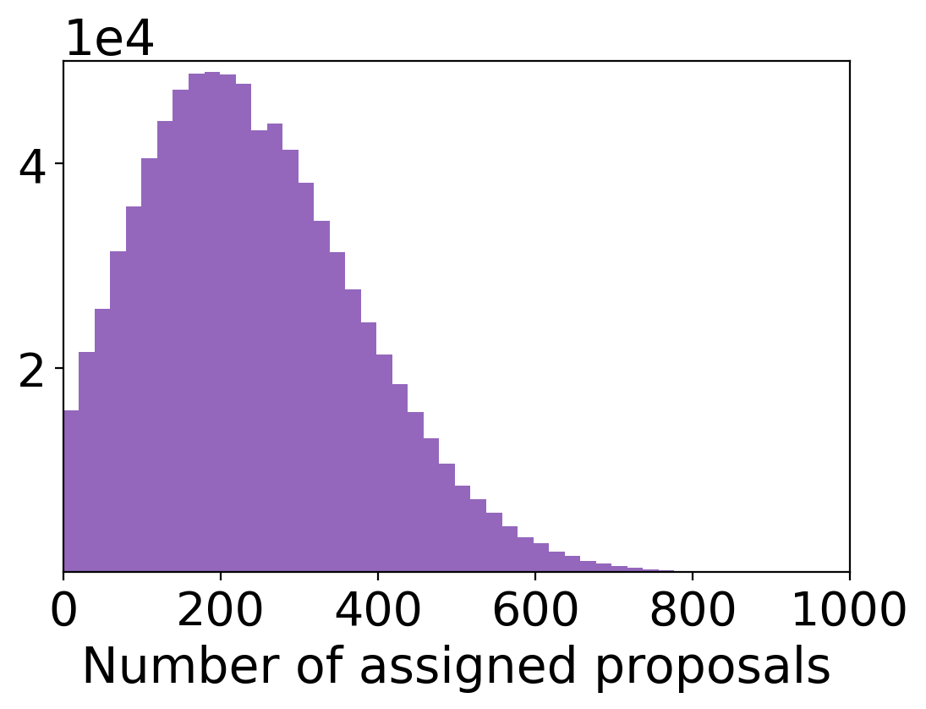}
\label{fig:a}
% \caption{}
}
\hfill
\subfigure[]{
\includegraphics[height=.22\columnwidth]{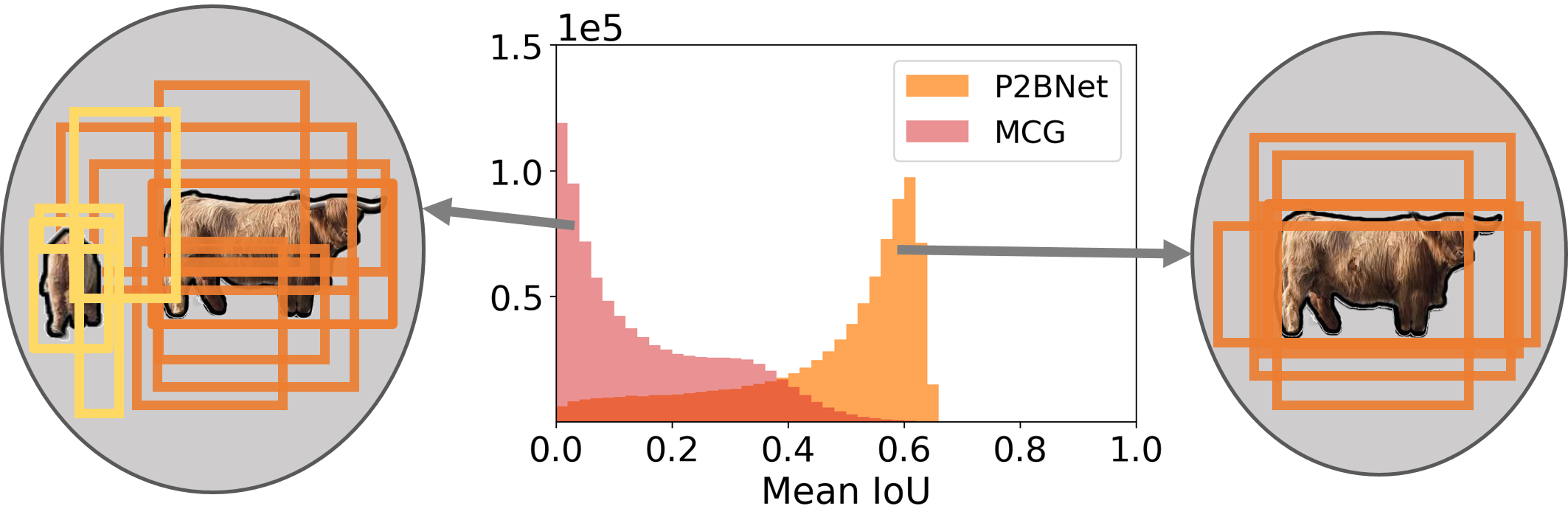}
% \caption{Chairs with different shapes.}
\label{fig:b}
}
%\vspace{-0.4cm}
\caption{(a) The number of assigned proposal boxes per object produced by MCG (\otsp~-based) is unbalanced, which is unfair for training. (b) Histogram of mIoU$_{prop}$
% $_{({\rm gt},{\rm bag})}$ 
for different proposal generation methods. mIoU$_{prop}$ denotes the mean IoU between proposal boxes and ground-truth for an object. Small mIoU$_{prop}$ in MCG brings semantic confusion. Whereas for our P2BNet with refinement, large mIoU$_{prop}$ is beneficial for optimization. Statistics are on COCO-17 training set, and both figures have 50 bins.}
%\vspace{-0.5cm}
\label{fig:meaniou_unbalance}
\end{figure}

Recently, point-based annotations are widely used in many tasks including object detection~\cite{Click,UFO2} and localization~\cite{CPR,DBLP:withoutboundingbox,p2pnet}, instance segmentation~\cite{cheng2021pointly}, and action localization~\cite{Lee_2021_ICCV}. However, the performance gap between point supervised detection methods~\cite{Click,UFO2} and bounding box supervised detectors remain large. Although it is understandable that location information provided by bounding boxes is richer than the points, we argue that this is not the only reason. We believe most PSOD methods do not utilize the full potential of point-based annotations. Previous works use off-the-shelf proposal (\otsp) methods (\eg, Selective Search~\cite{DBLP:selectivesearch}, MCG~\cite{DBLP:MCG}, and EdgeBox~\cite{DBLP:EdgeBox}) to obtain proposals for constructing bags.
% construct image-level bags
Despite the wide adaptation of these \otsp-based methods in weakly supervised detectors, they suffer from the following problems in Fig.~\ref{fig:motivation}: 1) There are too many background proposals in the bags. \otsp~methods generate too many proposal boxes that do not have any intersection with any of the foreground objects;
2) Positive proposals per object are unbalanced. The positive proposals per object produced by MCG on the COCO-17 training set are shown in Fig.~\ref{fig:meaniou_unbalance}(a), which is clearly off-balance; 3) Majority of the proposals in bags have very low IoU indicating low-quality proposals (Fig.~\ref{fig:meaniou_unbalance}(b)). 
Also, as the previous PSOD methods only construct image-level bags, they can not utilize the point annotations during MIL training leading to a mixture of different objects in the same bag.
All these problems limit the overall quality of the constructed bags, which contributes to the poor performance of the model.

% causing intra-class interference;

In this paper, we propose P2BNet as an alternative to the OTSP methods for generating high-quality object proposals. The number of proposals generated by P2BNet is balanced for each object, and they cover varied scales and aspect ratios. Additionally, the proposal bags are instance-level instead of image-level. This preserves the exclusivity of objects for a given proposal bag which is very helpful during MIL training. To further improve the quality of the bag, a coarse-to-fine procedure is designed in a cascade fashion in P2BNet. The refinement stage consists of two parts, the coarse pseudo-box prediction (CBP) and the precise pseudo-box refinement (PBR). The CBP stage predicts the coarse scale (width and height) of objects, whereas the PBR stage iteratively finetunes the scale and position. Our P2BNet generates high-quality, balanced proposal bags and ensures the contribution of point annotations in all stages (before, during, and after MIL training). The detailed experiments on COCO suggest the effectiveness and robustness of our model outperforming the previous point-based detectors by a large margin. Our main contributions are as follows:
%\vspace{-1.4mm}
\begin{itemize}
    \item[—]P2BNet, a generative and \otsp-free network, is designed for predicting pseudo boxes.
It generates inter-objects balanced instance-level bags and is beneficial for better optimization of MIL training. In addition, P2BNet is much more time-efficient than the \otsp-base methods. 

    \item[—]A coarse-to-fine fashion in P2BNet with CBP and PBR stage is proposed for higher-quality proposal bags and better prediction.
    \item[—]The detection performance of our proposed P2BNet-FR framework with P2BNet under single quasi-center point supervision improves the mean average precision (AP) of the previous best PSOD method by more than 50\% (relative) on COCO and bridges the gap between bounding box supervised detectors achieving comparable performance on AP$_{50}$.
\end{itemize}

%\vspace{-0.4cm}
\section{Related Work}
\label{RelatedWork}
%\vspace{-0.3cm}
In this section, we briefly discuss the research status of box-supervised, image-level and point-level supervised object detection. 

%\vspace{-0.4cm}
\subsection{Box-Supervised Object Detection}
%\vspace{-0.2cm}
Box-supervised object detection \cite{fastrcnn,FasterRCNN,DBLP:yolo,DBLP:SSD,DBLP:retinanet_focalloss,DBLP:DETR,DBLP:sparsercnn,scalematch} is a traditional object detection paradigm that gives the network a specific category and box information. 
% A series of box-supervised object detectors come with the development of the deep convolution neural network. 
One-stage detectors based on sliding-window, like YOLO~\cite{DBLP:yolo}, SSD~\cite{DBLP:SSD}, and RetinaNet~\cite{DBLP:retinanet_focalloss}, predict classification and bounding-box regression through setting anchors. Two-stage detectors predict proposal boxes through \otsp~methods (like selective search~\cite{DBLP:selectivesearch} in Fast R-CNN~\cite{fastrcnn}) or deep networks (like RPN in Faster R-CNN~\cite{FasterRCNN}) and conduct classification and bounding-box regression with filtered proposal boxes sparsely. 
% Motivated by transformer in natural language processing, 
Transformer-based detectors (DETR~\cite{DBLP:DETR}, Deformable-DETR~\cite{DBLP:DeformableDETR}, and Swin-Transformer~\cite{DBLP:swin-transformer}) come, utilizing global information for better representation. Sparse R-CNN~\cite{DBLP:sparsercnn} combines the advantages of transformer and CNN to a sparse detector.~\cite{DBLP:r3det,DBLP:roitransformer,DBLP:beyond_bbox} study on oriented object detection in aerial scenario. However, box-level annotation requires high costs.

%\vspace{-0.4cm}
\subsection{Image-Supervised Object Detection}
%\vspace{-0.1cm}
Image-supervised object detection~\cite{DBLP:wsddn,DBLP:oicr,DBLP:pcl,DBLP:slv,DBLP:MELM,DBLP:cam,DBLP:wccn,DBLP:Acol,DBLP:weaklysurvey,DBLP:journals/tip/MengZYZZW22,DBLP:uwsod} is the traditional field in WSOD. The traditional image-supervised WSOD methods can be divided into two styles: MIL-based~\cite{DBLP:wsddn,DBLP:oicr,DBLP:pcl,DBLP:slv,DBLP:MELM}, and CAM-based~\cite{DBLP:cam,DBLP:wccn,DBLP:Acol}.

In MIL-based methods, a bag is positively labelled if it contains at least one positive instance; otherwise, it is negative. The objective of MIL is to select positive instances from a positive bag. WSDDN~\cite{DBLP:wsddn} introduced MIL into WSOD with a representative two-stream weakly supervised deep detection network that can classify positive proposals. OICR~\cite{DBLP:oicr} introduces iterative fashion into WSOD and attempts to find the whole part instead of a discriminative part. PCL~\cite{DBLP:pcl} develops the proposal cluster learning and uses the proposal clusters as supervision to indicate the rough locations where objects most likely appear. Subsequently, SLV~\cite{DBLP:slv} brings in spatial likelihood voting to replace the max score proposal, further looking for the whole context of objects. Our paper produces the anchor-like~\cite{DBLP:uwsod,FasterRCNN} proposals around the point annotation as a bag and uses instance-level MIL to train the classifier. 
% We utilize a top-k average weighting policy instead of a max score proposal to find a similarly simple way of SLV. 
It moves the fixed pre-generated proposals (\eg OICR, PCL and UWSOD~\cite{DBLP:uwsod}) to achieve the coarse to fine purpose.

In CAM-based methods, the main idea is to produce the class activation maps (CAM)~\cite{DBLP:cam}, use threshold to choose a high score region, and find the smallest circumscribed rectangle of the largest general domain. WCCN~\cite{DBLP:wccn} uses a three-stage cascade structure. The first stage produces the class activation maps and obtains the initial proposals, the second stage is a segmentation network for refining object localization, and the last stage is a MIL stage outputting the results. Acol~\cite{DBLP:Acol} introduces two parallel-classifiers for object localization using adversarial complementary learning to alleviate the discriminative region.

%\vspace{-0.4cm}
\subsection{Point-Supervised Object Detection}
%\vspace{-0.1cm}
% %\vspace{-0.3cm}
Point-level annotation is a fairly recent innovation. The average time for annotating a single point is about 1.87s per image, close to image-level annotation (1.5 s/image) and much lower than that for bounding box(34.5 s/image). The statistics~\cite{DBLP:C-WSL,Click} are performed on VOC~\cite{DBLP:VOC}, which can be analogized to COCO~\cite{coco}.

\cite{Click} introduces center-click annotation to replace box supervision and estimates scale with the error between two times of center-click. \cite{UFO2} designs a network compatible with various supervision forms like tags, points, scribbles, and boxes annotation. However, these frameworks are based on \otsp~methods and are not specially designed for point annotation. 
Therefore, the performance is limited and performs poorly in complex scenarios like the COCO~\cite{coco} dataset. We introduce a new framework with P2BNet which is free of \otsp~methods. 

%\vspace{-0.3cm}
\section{Point-to-Box Network}
%\vspace{-0.2cm}
The P2BNet-FR framework consists of Point-to-Box Network (P2BNet) and Faster R-CNN (FR). P2BNet predicts pseudo boxes with point annotations to train the detector. We use standard settings for Faster R-CNN without any bells and whistles. Hence, we go over the proposed P2BNet in detail in this section.

\begin{figure}[t]
    \centering
    \includegraphics[width=1.\linewidth]{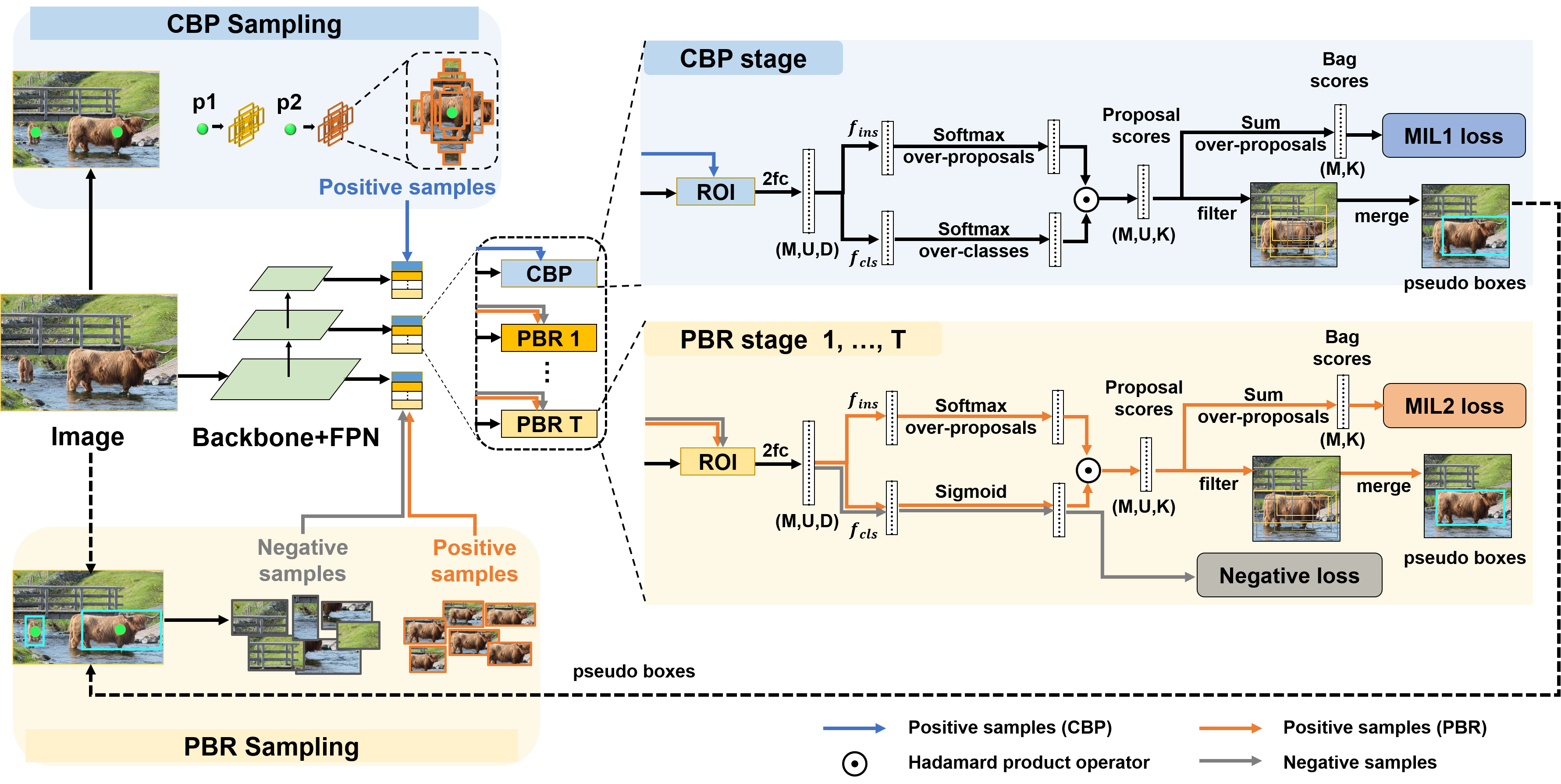}
    %\vspace{-0.8cm}
    \caption{The architecture of P2BNet. Firstly, to predict coarse pseudo boxes in CBP stage, proposal bags are fixedly sampled around point annotations for classifier training. Then, to predict refined pseudo boxes in PBR stage, high-quality proposal bags and negative proposals are sampled with coarse pseudo boxes for training. Finally, the pseudo boxes generated by the trained P2BNet serve as supervision for the training the classic detector. (Best viewed in color.)}
    \label{fig:framework}
%\vspace{-0.5cm}
\end{figure}

% The main purpose of P2BNet is to predict the box representation of the object with quasi-center point annotation in training set. The main procedure of P2BNet is proposal bag generation, coarse pseudo box prediction (CBP) and precise pseudo box refinement (PBR). 
The architecture of P2BNet is shown in Fig.~\ref{fig:framework}, which includes the coarse pseudo box prediction (CBP) stage and the pseudo box refinement (PBR) stage. 
The CBP stage predicts the coarse scale (width and height) of objects, whereas the PBR stage iteratively finetunes the scale and position.
The overall loss function of P2BNet is the summation of the losses of these two stages, \ie,
% The overall objective function of P2BNet with $T$ stages of the PBR module is defined in Eq.~\ref{Eq:P2B_loss_all}. It will be introduced in detail in the following sections.
%\vspace{-0.3cm}

\begin{equation}%\small
\begin{aligned}
\mathcal{L}_{p2b}= \mathcal{L}_{cbp} + \sum\limits_{t=1}^{T} \mathcal{L}_{pbr}^{(t)},
\label{Eq:P2B_loss_all}
% %\vspace{-0.2cm}
\end{aligned}
\end{equation}
where PBR includes $T$ iterations, and $\mathcal{L}_{pbr}^{(t)}$ is the loss of $t$-th iteration.
%\vspace{-0.3cm}

\begin{figure}[t]
    \centering
    \includegraphics[width=0.99\linewidth]{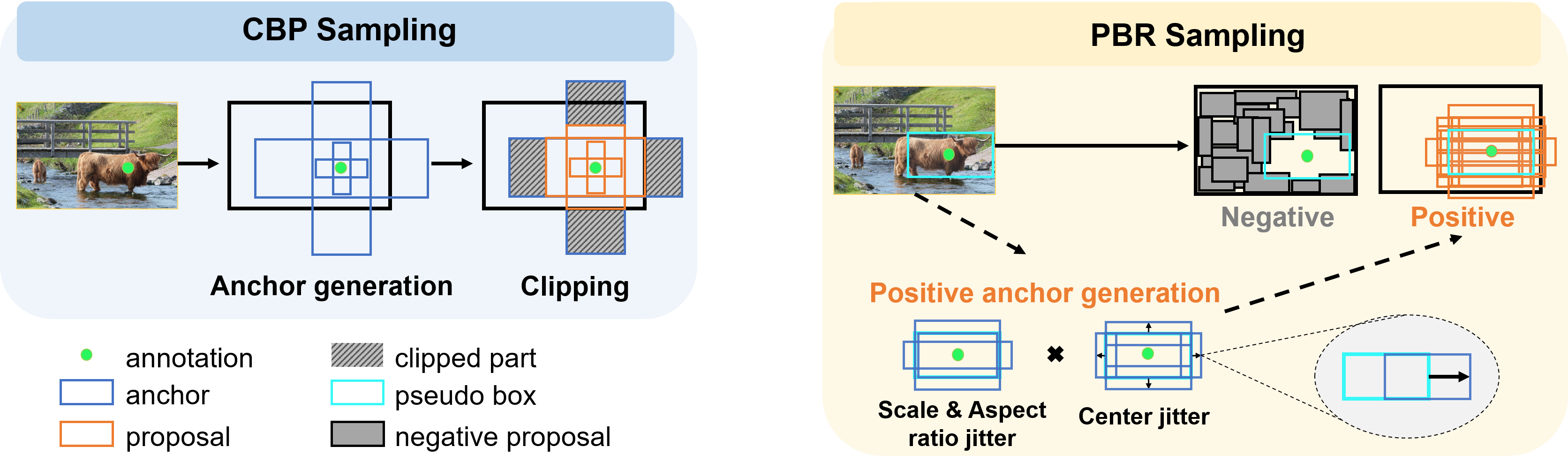}
    %\vspace{-0.2cm}
    \caption{Details of sampling strategies in the CBP stage and the PBR stage. The arrows in PBR sampling mean the offset of center jitter. Samples are obtained through center jitter following scale and aspect ratio jatter in PBR sampling.}
    \label{fig:sample_cbp_pbr}
%\vspace{-0.55cm}
\end{figure}

\subsection{Coarse Pseudo Box Prediction}
%\vspace{-0.2cm}
In the CBP stage, firstly, proposal boxes of different widths and heights are generated in an anchor-style for each object, taking the annotated point as the box center. Secondly, features of the sampled proposals are extracted to train a MIL classifier for selecting the best fitted proposal of objects.
% Finally, the top-$k$ (for stability) proposals with highest scores are weighted to predict the coarse pseudo box for each object.
Finally, the top-$k$ merging policy are utilized to estimate coarse pseudo boxes.

\textbf{CBP Sampling}: fixed sampling around the annotated point.
With the point annotation $p=(p_x,p_y)$ as the center, $s$ as the size, and $v$ to adjust the aspect ratio, the proposal box $b=(b_x, b_y, b_w, b_h)$ is generated, \ie~$b=(p_x, p_y, v\cdot s, \frac{1}{v} \cdot s)$. The schematic diagram of proposal box sampling is shown in Fig.~\ref{fig:sample_cbp_pbr} (Left). 
By adjusting $s$ and $v$, each point annotation $p_j$ generates a bag of proposal boxes with different scales and aspect ratios, denoted by $\mathcal{B}_j$ ($j \in \{1,2,\dots,M \}$, where $M$ is the amount of objects). The details of the settings of $s$ and $v$ are given in supplemental.
All proposal bags are utilized for training the MIL classifier in the CBP module with the category labels of points as supervision.

There is a minor issue that oversized $s$ may lead most of $b$ outside the image and introduce too many meaningless padding values. In this case, we clip $b$ to guarantee that it is inside the image (see Fig.~\ref{fig:sample_cbp_pbr} (Left)), \ie,
\begin{equation}%\small
\setlength\abovedisplayskip{5pt}%shrink space
\setlength\belowdisplayskip{5pt}
b = \bigg(p_x, p_y, min(v\cdot s,2(p_x-0),2(W-p_x)),
min(\frac{1}{v}\cdot s,2(p_y-0),2(H-p_y))\bigg),
\end{equation}where $W$ and $H$ denote the image size. $(p_x-0)$ and $(W-p_x)$ are the distances from the center to the left and right edges of the image, respectively. 
% We choose smaller as half the box width if the box exceeds the image size. The same goes for the box height. 

\textbf{CBP Module.}  
For a proposal bag $\mathcal{B}_j$, features $\mathbf{F}_j\in \mathbb{R}^{U \times D}$ are extracted through $7 \times 7$ RoIAlign~\cite{DBLP:MASK-RCNN} and two fully connected (fc) layers, where $U$ is the number of proposals in $\mathcal{B}_j$, and $D$ is the feature dimension. We refer to WSDDN~\cite{DBLP:wsddn} and design a two-stream structure as a MIL classifier to find the best bounding box region to represent the object. Specifically, applying the classification branch $f_{cls}$ to $\mathbf{F}_j$ yields $\mathbf{O}_{j}^{cls}\in \mathbb{R}^{U \times K}$, which is then passed through the activation function to obtain the classification score $\mathbf{S}^{cls}_j\in \mathbb{R}^{U \times K}$, where $K$ represents the number of instance categories. Likewise, instance score $\mathbf{S}^{ins}_j\in \mathbb{R}^{U \times K}$ is obtained through instance selection branch $f_{ins}$ and activation function, \ie, 
% %\vspace{-0.2cm}
\begin{equation}%\small
\setlength\abovedisplayskip{2pt}%shrink space
\setlength\belowdisplayskip{0pt}
% \begin{aligned}
\mathbf{O}^{cls}_j = f_{cls}(\mathbf{F}_j), \quad [\mathbf{S}^{cls}_j]_{uk} = e^{[\mathbf{O}^{cls}_j]_{uk}}\big/\sum\limits_{i=1}^{K} e^{[\mathbf{O}^{cls}_j]_{ui}}; \label{Eq:S_cls_branch}
\end{equation}
\begin{equation}%\small
\setlength\abovedisplayskip{0pt}
\mathbf{O}^{ins}_j = f_{ins}(\mathbf{F}_j), \quad [\mathbf{S}^{ins}_j]_{uk} = e^{[\mathbf{O}^{ins}_j]_{uk}}\big/\sum\limits_{i=1}^U e^{[\mathbf{O}^{ins}_j]_{ik}},
\label{Eq:S_ins_branch}
% \end{aligned}
\end{equation}
\noindent where $[\cdot]_{uk}$ denotes the value at row $u$ and column $k$ in the matrix. The proposal score $\mathbf{S}_j$ is obtained by computing the Hadamard product of the classification score and the instance score, and the bag score $\widehat{\mathbf{S}}_j$ is obtained by the summation of the proposal scores of $U$ proposal boxes, \ie,
%\vspace{-0.3cm}
\begin{equation}%\small
\begin{aligned}
\setlength\abovedisplayskip{1pt}%shrink space
\setlength\belowdisplayskip{1pt}
\mathbf{S}_j=\mathbf{S}^{cls}_j \odot \mathbf{S}^{ins}_j \in \mathbb{R}^{U \times K}, \quad
\widehat{\mathbf{S}}_j & = \sum\limits_{u=1}^{U} [\mathbf{S}_j]_u \in \mathbb{R}^{K}.
    %   & = \sum_{p\in B} ({e^{O^{ins}_p}}/{\sum_{p' \in B} e^{O^{ins}_{p'}}}) \cdot S_p\\
% %\vspace{-0.5cm}
\label{Eq:S_bag2}
% %\vspace{-0.9cm}
\end{aligned}
\end{equation}
% %\vspace{-0.1cm}
\noindent $\widehat{\mathbf{S}}_j$ can be seen as the weighted summation of the classification score $[\mathbf{S}^{cls}_j]_u$ by the corresponding selection score $[\mathbf{S}^{ins}_j]_u$.

\textbf{CBP Loss.} 
The MIL loss in the CBP module (termed $\mathcal{L}_{mil1}$ to distinguish it from the MIL loss in PBR) uses the form of cross-entropy loss, defined as: 
\begin{equation}%\small
\setlength\abovedisplayskip{3pt}%shrink space
\setlength\belowdisplayskip{3pt}
% \begin{aligned}
\mathcal{L}_{cbp} = \alpha_{mil1}\mathcal{L}_{mil1} =-\frac{\alpha_{mil1}}{M}\sum\limits_{j=1}^{M} \sum\limits_{k=1}^{K} [\mathbf{c}_j]_k \log([\widehat{\mathbf{S}}_j]_k) + (1-[\mathbf{c}_j]_k)  \log(1-[\widehat{\mathbf{S}}_j]_k),
\label{Eq:bce loss}
% \end{aligned} 
\end{equation}where $\mathbf{c}_{j} \in \{0, 1\}^{K}$ is the one-hot category label, $\alpha_{mil1}$ is 0.25.
The CBP loss is to make each proposal correctly predict the category and instance it belongs to. 
% The CBP loss is designed to enable CBP module justify whether the proposal box in $\mathcal{B}_j$ are in the same category with $p_j$ and belong to $p_j$.

Finally, the top-$k$ boxes with the highest proposal scores $\mathbf{S}_j$ of each object are weighted to obtain coarse pseudo boxes for the following PBR sampling. 
%\vspace{-0.4cm}

\subsection{Pseudo Box Refinement}
%\vspace{-0.1cm}
% \subsection{Coarse-point based pseudo box Estimation}
The PBR stage aims to finetune the position, width and height of pseudo boxes, and it can be performed iteratively in a cascaded fashion for better performance. By adjusting the height and width of the pseudo box obtained in the previous stage (or iteration) in a small span while jittering its center position, finer proposal boxes are generated as positive examples for module training. Further, because the positive proposal bags are generated in the local region, negative samples can be sampled far from the proposal bags to suppress the background. The PBR module also weights the top-$k$ proposals with the highest predicted scores to obtain the refined pseudo boxes, which are the final output of P2BNet.

\textbf{PBR Sampling}. adaptive sampling around estimated boxes.
% In PRB sampling, precise proposal boxes are generated from coarse pseudo boxes obtained in the CBP stage, 
As shown in Fig.~\ref{fig:sample_cbp_pbr} (Right), for each coarse pseudo box $b^*=(b^*_x,b^*_y,b^*_w,b^*_h)$ obtained in the previous stage (or iteration), we adjust its scale and aspect ratio with $s$ and $v$, and jitter its postion with $o_x, o_y$ to obtain the finer proposal $b=(b_x, b_y, b_w, b_h)$:
\begin{equation}%\small
\setlength\abovedisplayskip{2pt}%shrink space
\setlength\belowdisplayskip{0pt}
% \begin{aligned}
b_w = v \cdot s \cdot b^*_w, \quad b_h = \frac{1}{v} \cdot s \cdot b^*_h, 
\setlength\abovedisplayskip{0pt}
\label{Eq: PRB sampling}
% \end{aligned}
\end{equation}
\begin{equation}%\small
\setlength\abovedisplayskip{2pt}%shrink space
\setlength\belowdisplayskip{0pt}
% \begin{aligned}
\setlength\abovedisplayskip{0pt}
b_x = b^*_x + b_w \cdot o_x, \quad b_y = b^*_y + b_h \cdot o_y.
\label{Eq: PRB sampling 2}
% \end{aligned}
\end{equation}
% (b^*_x+ b^*_w \cdot s \cdot v, b^*_y + o \cdot b^*_h \cdot \frac{1}{v}, v \cdot b^*_w, \frac{1}{v} \cdot b^*_h)$. 
These finer proposals are used as positive proposal bag $\mathcal{B}_j$ to train PBR module.

\begin{figure}[t]
    \centering
    \includegraphics[width=1.\linewidth]{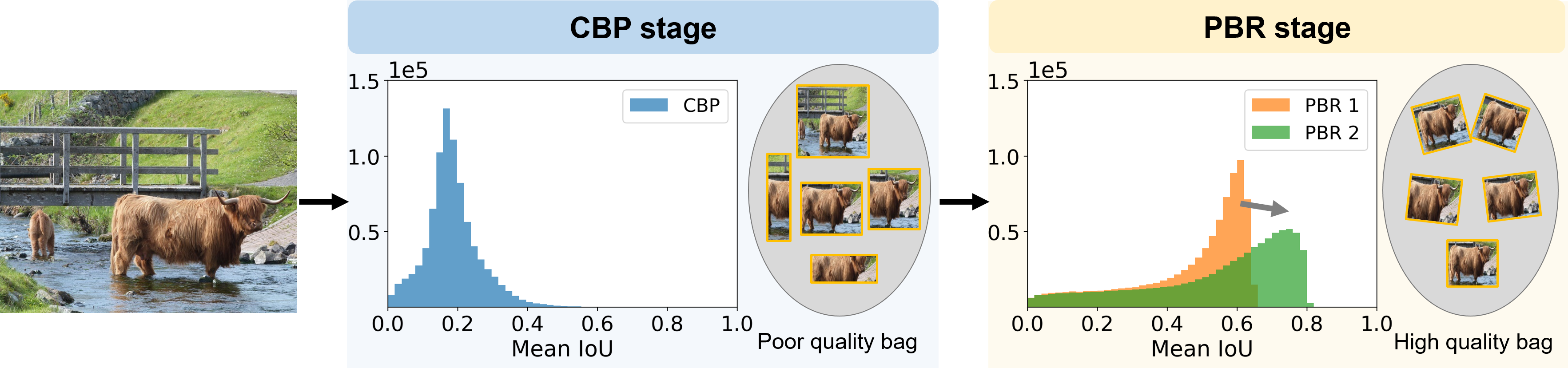}
    %\vspace{-0.8cm}
     \caption{The progression of the mIoU$_{prop}$ during refinement. By statistics, the mIoU$_{pred}$ is gradually increasing in the PBR stage, indicating that the quality of the proposal bag improves in iterative refinement.}
    \label{fig:coarse to fine}
%\vspace{-0.5cm}
\end{figure}

Furthermore, to better suppress the background, negative samples are introduced in the PBR sampling. 
We randomly sample many proposal boxes, which have small IoU (by default set as smaller than 0.3) with all positive proposals in all bags, to compose the negative sample set $\mathcal{N}$ for the PBR module. Through sampling proposal boxes by pseudo box distribution, high-quality proposal boxes are obtained for better optimization (shown in Fig.~\ref{fig:coarse to fine}).

\textbf{PBR Module.} 
The PBR module has a similar structure to the CBP module. It shares the backbone network and two fully connected layers with CBP, and also has a classification branch $f_{cls}$ and an instance selection branch $f_{ins}$. Note that $f_{cls}$ and $f_{ins}$ do not share parameters between different stages and iterations.
% The structure of PBR module is the same as CBP module. It shares the same backbone and two fc layers with CBP and has additional classification branch and instance selection branch. 
For instance selection branch, we adopt the same structure as the CBP module, and utilize Eq.~\ref{Eq:S_ins_branch} to predict the instance score $\mathbf{S}^{ins}_j $ for the proposal bag $\mathcal{B}_j$. Differently, the classification branch uses the $sigmoid$ activation function $\sigma(x)$ to predict the classification score $\mathbf{S}^{cls}_j $ , \ie,
\begin{equation}%\small
\begin{aligned}
\sigma(x) = 1/(1+e^{-x}), \quad
\mathbf{S}^{cls}_j = \sigma(f_{cls}(\mathbf{F}_j)) \in \mathbb{R}^{U \times K}.
\label{Eq:cls_pbr}
\end{aligned}    
\end{equation}
% %\vspace{-0.1cm}
This form makes it possible to perform multi-label classification, which can distinguish overlapping proposal boxes from different objects. According to the form of Eq.~\ref{Eq:S_bag2}, bag score $\mathbf{\widehat{S}}^*_j$ is calculated using $\mathbf{S}^{cls}_j$ and $\mathbf{S}^{ins}_j$ of the current stage. 

For the negative sample set $\mathcal{N}$, we calculate its classification score as:
\begin{equation}%\small
\begin{aligned}
\mathbf{S}^{cls}_{neg} = \sigma(f_{cls}(\mathbf{F}_{neg})) \in \mathbb{R}^{\left|\mathcal{N}\right|\times K}.
\label{Eq:bg loss}
\end{aligned}    
\end{equation}

\textbf{PBR Loss.} The PBR loss consists of MIL loss $\mathcal{L}_{mil2}$ for positive bags and negative loss $\mathcal{L}_{neg}$ for negative samples, \ie,
\begin{equation}%\small
\begin{aligned}
% L_{CPR} = & \alpha_B L_B + \alpha_{ann} L_{ann} +\alpha_{neg} L_{neg}
\mathcal{L}_{pbr} = \alpha_{mil2} \mathcal{L}_{mil2}  +\alpha_{neg} \mathcal{L}_{neg},
\label{Eq:PBR basic loss}
\end{aligned}
\end{equation}
\noindent where $\alpha_{mil2}=0.25$ and  $\alpha_{neg}=0.75$ are the settings in this paper.

\textbf{1) MIL Loss.} The MIL loss $\mathcal{L}_{mil2}$ in the PBR stage is defined as:
%\vspace{-0.2cm}
\begin{equation}%\small
\begin{aligned}
\setlength\abovedisplayskip{1pt}%shrink space
\setlength\belowdisplayskip{1pt}
{\rm FL}(\mathbf{\zeta}, \mathbf{\tau}) =& - \sum\limits_{k=1}^{K} [\mathbf{\tau}]_k  (1 - [\mathbf{\zeta}]_k)^{\gamma} \log([\mathbf{\zeta}]_k) + (1-[\mathbf{\tau}]_k) ([\mathbf{\zeta}]_k)^{\gamma} \log(1-[\mathbf{\zeta}]_k),
\label{Eq:focal loss}
\end{aligned}
\end{equation}
%\vspace{-0.4cm}
\begin{equation}%\small
\begin{aligned}
\setlength\abovedisplayskip{0pt}%shrink space
\setlength\belowdisplayskip{1pt}
\mathcal{L}_{mil2}  = \frac{1}{M}\sum\limits_{j=1}^{M} \left< \mathbf{c}^{\mathrm{T}}_j, \mathbf{\widehat{S}}^*_j \right> \cdot {\rm FL}(\mathbf{\widehat{S}}_j, \mathbf{c}_j),
\label{Eq:L_{bag}}
\end{aligned}
\end{equation}
% %\vspace{-0.2cm}
\noindent where ${\rm FL}(\mathbf{\zeta}, \mathbf{\tau})$ is the focal loss~\cite{DBLP:retinanet_focalloss}, and $\gamma$ is set as 2 following~\cite{DBLP:retinanet_focalloss}. 
% $\mathbf{c}_j$ is the one-hot category label of $p_j$.
$\mathbf{\widehat{S}}^*_j$ represents the bag score of the last PBR iteration (for the first iteration of PBR, using the bag score in CBP). ${\small{\left< \mathbf{c}^{\mathrm{T}}_j, \mathbf{\widehat{S}}^*_j \right>}}$ represents the inner product of the two vectors, which means the predicted bag score of the previous stage or iteration on ground-truth category. Score is used to weight the $\rm FL$ of each object for stable training.

\textbf{2) Negative Loss.} Conventional MIL treats proposal boxes belonging to other categories as negative samples. In order to further suppress the backgrounds, we sample more negative samples in the PBR stage and introduce the negative loss ($\gamma$ is also set to 2 following $\rm FL$), \ie,
% However, positive proposal bag is sampled in local region in the PBR stage, more negative samples are required to be suppressed. Hence, we introduce the Negative loss $L_{neg}$ in Eq.\ref{Eq:neg_pbr}.
%\vspace{-0.6cm}

\begin{equation}%\small
\begin{aligned}
\setlength\abovedisplayskip{1pt}%shrink space
\setlength\belowdisplayskip{1pt}
\beta =\frac{1}{M}\sum\limits_{j=1}^{M} \left< \mathbf{c}^{\mathrm{T}}_j, \mathbf{\widehat{S}}^*_j \right>, \quad \mathcal{L}_{neg} = - \frac{1}{\left|\mathcal{N}\right|}\sum\limits_{\mathcal{N}}\sum\limits_{k=1}^{K} \beta \cdot ([\mathbf{S}^{cls}_{neg}]_k)^{\gamma} \log(1-[\mathbf{S}^{cls}_{neg}]_k).
\label{Eq:neg_pbr}
% %\vspace{-0.6cm}
\end{aligned}
\end{equation}

%\vspace{-0.6cm}
\section{Experiments}
%\vspace{-0.1cm}
% \subsection{Spotting-point based pseudo box Estimation}
% #CPR+Coarse-point based pseudo box Estimation
\subsection{Experiment Settings}
% %\vspace{-0.2cm}
\textbf{Datasets and Evaluate Metrics.} For experiments, we use the public
available MS COCO~\cite{coco} dataset. COCO has 80 different categories and two versions. COCO-14 has 80K training and 40K validation images whereas COCO-17 has 118K training and 5K validation images. Since the ground truth on the test set is not released, we train our model on the training set and evaluate it on the validation set
% \textbf{Pascal VOC} is Pascal VOC 2007 and Pascal VOC 2012, which both contain 20 object categories. In VOC 2007, models are trained on 5011(2501+2510) images of the trainval set and evaluated on 4951 images of the test set. In VOC 2012, 11540 (5717+5823) images are for the trainval set, and 10991 images are for the test set.
reporting AP$_{50}$ and AP (averaged over IoU thresholds in $[0.5:0.05:0.95]$) on COCO. The mIoU$_{pred}$ is calculated by the mean IoU between predicted pseudo boxes and their corresponding ground-truth bounding-boxes of all objects in the training set. It can directly evaluate the ability of P2BNet to transform annotated points into accurate pseudo boxes.

\textbf{Implementation Details.} Our codes of P2BNet-FR are based on MMDetection~\cite{mmdetection}. 
%  Same to the default setting of the object detection on COCO, 
The stochastic gradient descent (SGD~\cite{DBLP:series/lncs/Bottou12}) algorithm is used to optimize in 1$\times$ training schedule. The learning rate is set to 0.02 and decays by 0.1 at the 8-th and 11-th epochs, respectively. 
% In P2BNet, we use multi-scale images (480, 576, 688, 864, 1000, 1200) for a shorter edge in the training and inference pseudo box in 1200 scale, unless otherwise specified.
In P2BNet, we use multi-scale (480, 576, 688, 864, 1000, 1200) as the short side to resize the image during training and single-scale (1200) during inference.
We choose the classic Faster R-CNN FPN~\cite{FasterRCNN,DBLP:FPN} (backbone is ResNet-50~\cite{DBLP:resnet}) as the detector with the default setting, and single-scale (800) images are used during training and inference. More details are included in the supplementary section.
% For CBP stage box generation, we set $S=\{4,8,16,32,64,128\}\cdot min(W,H)/100$ and $U=\{1,1.5,2,3,1/1.5,1/2,1/3\}$. For PBR stage box generation, we set $V=\{1, 1.2, 1.3, 0.8, 0.7\}$ and $O=\{(0.1,0.1),(0.1,-0.1),(-0.1,-0.1),(-0.1,0.1)\}$. [TODO] For negative sampling, we set the neg num = 500  [TODO].

\textbf{Quasi-Center Point Annotation.} We propose a quasi-center (QC) point annotation that is friendly for object detection tasks with a low cost. In practical scenarios, we ask annotators to annotate the object in the non-high limit center region with a loose rule. Since datasets in the experiment are already annotated with bounding boxes or masks, it is reasonable that the manually annotated points follow Gaussian distribution in the central region.
% according to the law of large numbers. 
% Considering that the annotated points must be in the central region of the object's bounding box or mask, w
We utilize Rectified Gaussian Distribution ($RG$) defined in \cite{CPR} with central ellipse constraints.
For a bounding box of $b=(b_x,b_y,b_w,b_h)$, its central ellipse can be defined as $Ellipse(\kappa)$, using $(b_x,b_y)$ as the ellipse center and $(\kappa \cdot b_w, \kappa \cdot b_h)$ as the two axes of the ellipse. 
In addition, in view of the fact that the absolute position offset for a large object is too large under the above rule, we limit the two axes to no longer than 96 pixels. 
If the object's mask $Mask$ overlaps with the central ellipse $Ellipse(\kappa)$, $V$ is used to denote the intersection. If there is no intersecting area, $V$ represents the entire $Mask$. 
When generated from bounding box annotations, the boxes are treated as masks. Then $RG$ is defined as,
\begin{equation}
RG(p;\mu,\sigma , \kappa)=\left\{
\begin{array}{rcl}
& \frac{Gauss(p;\mu,\sigma)}{\int_V Gauss(p;\mu,\sigma)dp}, & {p \in V}\\
& 0,  & {p \notin V}
\end{array} \right.
\end{equation}
\noindent where $\mu$ and $\sigma$ are mean and standard deviation of $RG$. $\kappa$ decides the $Ellipse(\kappa)$. In this paper, $RG(p;0, \frac{1}{4},\frac{1}{4})$ is chosen to generate the QC point annotations. 
% For statistics, 200 images were manually annotated by three annotators. The average annotation time per object is 4.27s for quasi-center strategy and 3.84s for the previous random annotation style\cite{DBLP:Points_As_Queries}. Considering the huge performance gain, we believe this is a fair trade-off.
%\vspace{-0.4cm}
% The box will be treated as a mask when generated from the bounding box annotation.

\begin{table}[tb!]
    \setlength{\tabcolsep}{3pt}
    \renewcommand\arraystretch{0.65}
    \centering
    \begin{tabular}{l|c|c|cc|cc}
    % \hline
    
    % \hline
    \specialrule{0.13em}{0pt}{1pt}  
     \multirow{2}{*}{Method} & \multirow{2}{*}{Backbone} & \multirow{2}{*}{Proposal}  & \multicolumn{2}{c|}{COCO-14} & \multicolumn{2}{c}{COCO-17}  \\
    \cline{4-7}
    &&& AP & AP$_{50}$ & AP & AP$_{50}$ \\
    % \hline
    \specialrule{0.08em}{0pt}{1pt}
    \multicolumn{5}{l}{\textbf{\emph{Box-supervised detectors}}} \\
    \specialrule{0.08em}{0pt}{0pt}
     Fast R-CNN~\cite{fastrcnn}  & VGG-16 &SS  & 18.9 & 38.6 &19.3&39.3\\
    Faster R-CNN~\cite{FasterRCNN} & VGG-16 & RPN  & 21.2 &41.5 &21.5 &42.1 \\
    FPN~\cite{mmdetection} &R-50&RPN &\textbf{35.5}&\textbf{56.7}&37.4&\textbf{58.1}\\
    RetinaNet~\cite{DBLP:retinanet_focalloss,mmdetection}  &R-50&-&34.3&53.3&36.5&55.4 \\
    Reppoint~\cite{DBLP:reppoint,mmdetection}  &R-50&-&-&-&37.0&56.7 \\
    Sparse R-CNN~\cite{DBLP:sparsercnn,mmdetection} &R-50&PP&-&-&\textbf{37.9}&56.0 \\
    \specialrule{0.08em}{0pt}{1pt}
    %  &WSDDN & VGG-16& SS& -&-&-&11.5\\
    %  &WCCN & VGG-16& SS & -&-&-&12.3\\
    % & \cite{GAN-base}+SSD &  VGG-16&- &-&-&-&13.6 \\
    \multicolumn{5}{l}{\textbf{\emph{Image-supervised detectors}}} \\
    \specialrule{0.08em}{0pt}{0pt}
    OICR+Fast~\cite{DBLP:oicr,fastrcnn} &  VGG-16 &SS&7.7&17.4&-&-\\
     PCL~\cite{DBLP:pcl} & VGG-16&SS& 8.5&19.4&-&-\\   
     PCL+Fast~\cite{DBLP:pcl,fastrcnn}    &VGG-16&SS & 9.2 &19.6&-&-\\ 
     MEFF+Fast~\cite{MEFF,fastrcnn}  &  VGG-16&SS& 8.9&19.3&-&-\\

     C-MIDN~\cite{C-MIDN} &VGG-16& SS& 9.6&21.4&-&-\\
    WSOD2~\cite{WSOD2} & VGG-16&SS& 10.8&22.7&-&-\\
    UFO$^{2*}$~\cite{UFO2} &VGG-16& MCG  &10.8 &23.1 &-&-\\
    GradingNet-C-MIL~\cite{DBLP:guidingnet} & VGG-16 & SS & 11.6 & 25.0 &-&-\\
     ICMWSD~\cite{ICMWSD} & VGG-16 & MCG  & 11.4 &24.3&-   &-\\
    %  &12.1&24.8 \\
       ICMWSD~\cite{ICMWSD}     &   R-50 & MCG   & 12.6 &26.1 &-   &-    \\
       ICMWSD~\cite{ICMWSD}   &   R-101 & MCG   & 13.0 &26.3 &-   &-    \\
     CASD~\cite{CASD} &  VGG-16 & SS  &12.8 & 26.4  &-   &-    \\
     CASD~\cite{CASD}    &     R-50 & SS   & \textbf{13.9} &\textbf{27.8} &-   &-    \\ 
     
    \specialrule{0.08em}{0pt}{1pt}
    \multicolumn{5}{l}{\textbf{\emph{Point-supervised detectors}}} \\
    \specialrule{0.08em}{0pt}{0pt}
     Click~\cite{Click}   & AlexNet&SS  & - &18.4 &-&- \\
     UFO$^2$~\cite{UFO2}      &VGG-16& MCG  &12.4 &27.0 &-&-\\
    UFO$^{2{\dagger}}$~\cite{UFO2}     &VGG-16& MCG    &12.8 &26.6 &13.2 &27.2\\ 
    %  &UFO$^2$\cite{UFO2}(we)     &VGG-16& SS  &13.3 &27.5 &-&-\\
    UFO$^{2{\ddagger}}$~\cite{UFO2}  &VGG-16& MCG  &12.7& 26.5 &13.5 &27.9 \\
    % (QC) &UFO$^2$\cite{UFO2}(we)    &R50-fpn& SS &-&- &13.1 &28.2\\
    UFO$^{2{\ddagger}}$~\cite{UFO2}   & R-50 & MCG &12.6&27.6&13.2 &28.9 \\    
    % (QC) &UFO$^2$\cite{UFO2}(we)   & COCO-17  &R50& SS   &12.4 &25.3 &-&-\\
    % \hline
     P2BNet-FR (Ours)   & R-50 & Free   &\textbf{19.4}&\textbf{43.5} & \textbf{22.1} & \textbf{47.3} \\
%   (QC) &P2BNet-FR(our)  &(R-101+R-50)-FPN & Free &-&-  & 21.85 & 47.48 \\
    \specialrule{0.13em}{0pt}{0pt}
    \end{tabular}
    %\vspace{1em}
    \caption{The performance comparison of box-supervised, image-supervised, and point-supervised detectors on COCO dataset. $^*$ means UFO$^2$ with image-level annotation. $^{\dagger}$ means the performance we reproduce with the original setting. $^{\ddagger}$ means we re-implement UFO$^2$ with our QC point annotation. The performance of P2BNet-FR, UFO$^2$, and the box-supervised detector is tested on a single scale dataset. Our P2BNet-FR is based on P2BNet with top-4 merging and one PBR stage. SS is selective search~\cite{DBLP:selectivesearch}, PP means proposal box defined in \cite{DBLP:sparsercnn}, and Free represents \otsp-free based method.}
    %\vspace{-3em}
    % In backbone part, R-50 is resnet\cite{DBLP:resnet}-50, VGG-16 is in \cite{DBLP:VGG}. In proposal part, SS is selective search\cite{DBLP:selectivesearch}, RPN is Region Proposal Network\cite{FasterRCNN}, PP is proposal box in \cite{DBLP:sparsercnn}, MCG is in \cite{DBLP:MCG}.
    % }
    \label{tab:coco-tmp}
\end{table}

\subsection{Performance Comparisons}
Unless otherwise specified, the default components of our P2BNet-FR framework are P2BNet and Faster R-CNN. We compare the P2BNet-FR with the existing PSOD methods while choosing the state-of-the-art UFO$^2$~\cite{UFO2} framework as the baseline for comprehensive comparisons.
% on multiple datasets and backbones.
In addition, to demonstrate the performance advantages of the PSOD methods, we compare them with the state-of-the-art WSOD methods. At the same time, we compare the performance of the box-supervised object detectors to reflect their performance upper bound.

\begin{figure}[t]
        \centering
        \includegraphics[width=1.\linewidth]{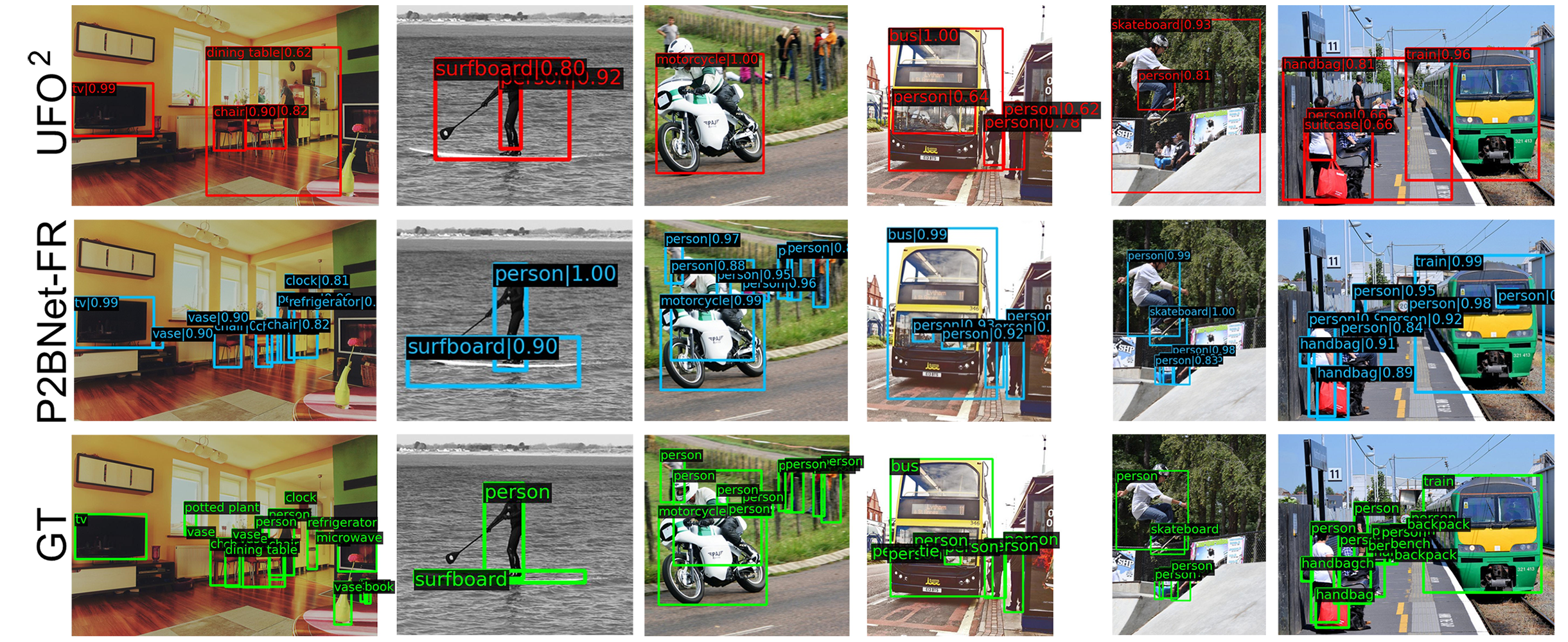}
        %\vspace{-1.0em}
    \caption{Visualization of detection results of P2BNet-FR and UFO$^2$. Our P2BNet-FR can distinguish dense objects and perform well in complex scene. (Best viewed in color.)}
    %\vspace{-0.6cm}
    \label{fig:Vis}
\end{figure}

\textbf{Comparison with PSOD Methods.}
We compare the existing PSOD methods Click~\cite{Click} and UFO$^2$~\cite{UFO2} on COCO, as shown in Tab.~\ref{tab:coco-tmp}. Both Click and UFO$^2$ utilize \otsp-based methods (SS~\cite{DBLP:selectivesearch} or MCG~\cite{DBLP:MCG}) to generate proposal boxes. Since the point annotation used by UFO$^2$ is different from the QC point proposed in this paper, for a fair comparison, we re-train UFO$^2$ on the public code with our QC point annotation. In addition, the previous methods are mainly based on VGG-16~\cite{DBLP:VGG} 
or AlexNet~\cite{DBLP:alexnet}.
For consistency, we extend the UFO$^2$ to the ResNet-50 FPN backbone and compare it with our framework.
In comparison with Click and UFO$^2$, our P2BNet-FR framework outperforms them by a large margin. On COCO-14, P2BNet-FR improves AP and AP$_{50}$ by 6.8 and 15.9, respectively. Also, our framework significantly outperforms state-of-the-art performance by 8.9 AP and 18.4 AP$_{50}$ on COCO-17. In Fig.~\ref{fig:Vis}, the visualization shows our P2BNet-FR makes full use of the precise location information of point annotation and can distinguish dense objects in complex scenes.

\textbf{Comparison with WSOD Methods.} 
We compare the proposed framework to the state-of-the-art WSOD methods on the COCO-14 in Tab.~\ref{tab:coco-tmp}. 
The performance of P2BNet-FR proves that compared with WSOD, PSOD significantly improves the detection performance with little increase in the annotation cost, showing that the PSOD task has great prospects for development.

\textbf{Comparison with Box-Supervised Methods.}
In order to verify the feasibility of P2BNet-FR in practical applications and show the upper bound under this supervised manner, we compare the box-supervised detector~\cite{FasterRCNN} in Tab.~\ref{tab:coco-tmp}. Under AP$_{50}$, P2BNet-FR-R50 (47.3 AP$_{50}$) is much closer to box-supervised detector FPN-R50 (58.1 AP$_{50}$) than previous WSOD and PSOD method. It shows that PSOD can be applied in industries that are less demanding on box quality and more inclined to find objects~\cite{DBLP:ANTI-UAV,DBLP:2ND_ANTI-UAV}, with greatly reduced annotation cost.

\begin{table}[tb!]
\centering
\begin{minipage}[htb]{1.0\linewidth} 
\centering
    \begin{minipage}[htb]{0.8\linewidth}
        \setlength{\tabcolsep}{3pt}
        \renewcommand\arraystretch{0.65}
        \centering
        \begin{tabular}{cc|ccc|ccc}
        \specialrule{0.13em}{0pt}{1pt}
            \multicolumn{2}{c|}{CBP stage} & \multicolumn{3}{c|}{PBR stage} & \multicolumn{3}{c}{Performance}\\
            \specialrule{0.08em}{0pt}{1pt}
            $\mathcal{L}_{pos}$ &
            $\mathcal{L}_{mil1}$ & $\mathcal{L}_{mil2}$ & $\mathcal{L}_{neg}$ & $\mathcal{L}_{pesudo}$ & mIoU$_{pred}$& AP & AP$_{50}$\\
            % \hline
            % \hline
            \specialrule{0.08em}{0pt}{1pt}
             \checkmark & & \multicolumn{3}{c|}{} &25.0 &2.9 &10.3 \\
            & \checkmark  & \multicolumn{3}{c|}{}& 50.2 &13.7 & 37.8\\
            \specialrule{0.08em}{0pt}{1pt} 
            % & \checkmark && & & 48.0 &13.3 &36.6\\
            & \checkmark &\checkmark &  &&52.0&12.7 &35.4\\
            &  \checkmark& \checkmark&\checkmark& &\textbf{57.4}&  \textbf{21.7}& \textbf{46.1}\\
            &\checkmark& \checkmark & \checkmark & \checkmark& 56.7 &18.5 & 44.1 \\ 
            \specialrule{0.13em}{0pt}{0pt}
        \end{tabular}
        %\vspace{-0.5em}
        \flushleft{\par{(a) The effectiveness of training loss in P2BNet: $\mathcal{L}_{mil1}$ in CBP stage, $\mathcal{L}_{mil2}$ and $\mathcal{L}_{neg}$ in PBR stage. $\mathcal{L}_{pos}$  and $\mathcal{L}_{pesudo}$ is for comparison.}}
        \label{tab:main loss}
    \end{minipage}
    \begin{minipage}[htb]{1.0\linewidth} %\vspace{3mm}
    \centering
         \begin{minipage}[htb]{0.45\linewidth}
            \setlength{\tabcolsep}{4pt}
            \renewcommand\arraystretch{0.65}
            \centering
            % \begin{spacing}{1.19}
            \begin{tabular}{c|ccc}
            \specialrule{0.13em}{0pt}{1pt}
            top-$k$ &mIoU$_{pred}$ &AP &AP$_{50}$\\
            \specialrule{0.08em}{0pt}{1pt}
                 1& 49.2 &12.2 &35.9 \\
                 3& 54.7 &21.3 & 46.6\\
                 4& \textbf{57.5} & \textbf{22.1} & \textbf{47.3}\\
                 7& 57.4 & 21.7 & 46.1\\ 
                 10& 57.1 & 21.5 & 46.0\\
            \specialrule{0.13em}{0pt}{0pt}
            \end{tabular}
            % \caption{(b) topk insensitive}
            %\vspace{-0.5em}
            \flushleft{\par{(b) The top-$k$ policy for box merging. $k$ is set the same for all stages.}}
            \label{tab:top k }
        \end{minipage}\quad
        \begin{minipage}[htb]{0.40\linewidth}
        \setlength{\tabcolsep}{4.3pt}
        \renewcommand\arraystretch{0.6}
            \centering
            \begin{tabular}{c|ccc}
            \specialrule{0.13em}{0pt}{1pt}
            $T$ &mIoU$_{pred}$& AP &AP$_{50}$\\
            \specialrule{0.08em}{0pt}{1pt}
                 0& 50.2  &13.7 & 37.8 \\
                 1& \textbf{57.4} & 21.7& \textbf{46.1} \\
                 2& 57.0 &\textbf{21.9} & \textbf{46.1} \\
                 3& 56.2 &  21.3  &   45.6 \\
            \specialrule{0.13em}{0pt}{0pt}
            \end{tabular}
            % \caption{number of refine stage}
            %\vspace{-0.5em}
            \flushleft{\par{(c) The number of iterations $T$ in the PBR stage. $T=0$ means only the CBP stage is conducted.}}
            \label{tab:num-refine}
        \end{minipage}
    \end{minipage}

\end{minipage}

    %\vspace{0.4em}
    \caption{Ablation study (Part I).}
    \label{tab: ablation stdudy i}
    %\vspace{-3em}
\end{table}

%\vspace{-0.4cm}
\subsection{Ablation Study}
%\vspace{-0.2cm}
In this section, all the ablation studies are conducted on the COCO-17 dataset. The top-$k$ setting is $k=7$ except for the box merging policy part in Tab.~\ref{tab: ablation stdudy i}(b) and different detectors part ($k=4$) in  Tab.~\ref{tab: ablation study ii}(d).

\noindent\textbf{Training Loss in P2BNet.} The ablation study of the training loss in P2BNet is shown in Tab.~\ref{tab: ablation stdudy i}(a). \textbf{1) CBP loss.} Only with $\mathcal{L}_{mil1}$ in the CBP stage, we can obtain 13.7 AP and 37.8 AP$_{50}$. For comparison, we conduct $\mathcal{L}_{pos}$, which views all the proposal boxes in the bag as positive samples. We find it hard to optimize, and the performance is bad, demonstrating the effectiveness of our proposed $\mathcal{L}_{mil1}$ for pseudo box prediction. Coarse proposal bags can cover most objects in high IoU, resulting in a low missing rate. However, the performance still has the potential to be refined because the scale and aspect ratio are coarse, and the center position needs adjustment. 
% Consequently, the PBR stage comes with its $\mathcal{L}_{mil2}$ and $\mathcal{L}_{neg}$.
\noindent\textbf{2) PBR loss.} 
With a refined sampling of proposal bag (shown in Fig.~\ref{fig:coarse to fine}), corresponding PBR loss is introduced. Only with $\mathcal{L}_{mil2}$, the performance is just 12.7 AP. 
The main reasons of performance degradation are error accumulation in a cascade fashion and lacking negative samples for focal loss. 
% Poor prediction becomes worse in cascade style. 
% Meanwhile, $Sigmoid$ activation function needs enough negative samples to suppress background. 
% Therefore, only using $\mathcal{L}_{mil2}$ will bring training failure and further lead to performance drop.
There are no explicit negative samples to suppress background for $Sigmoid$ activation function, negative sampling and negative loss $\mathcal{L}_{neg}$ is introduced.
% which brings the $\mathcal{L}_{mil2}$ for the focal loss style.  we conduct $\mathcal{L}_{neg}$ in PBR, 
Performance increases by 9.0 AP and 10.7 AP$_{50}$, indicating that it is essential and effectively improves the optimization. We also evaluate the mIoU$_{pred}$ to discuss the predicted pseudo box's quality. 
% In CBP stage with $\mathcal{L}_{mil1}$, the mIoU$_{pred}$ is 50.2. 
In the PBR stage with $\mathcal{L}_{mil2}$ and $\mathcal{L}_{neg}$, the mIoU increases from 50.2 to 57.4, suggesting better quality of the pseudo box. Motivated by \cite{CPR}, we conduct $\mathcal{L}_{pesudo}$, viewing pseudo boxes from the CBP stage as positive samples. However, the $\mathcal{L}_{pesudo}$ limits the refinement and the performance decreases. 
% We also analyze the influence of jitter strategy. 
In Tab.~\ref{tab: ablation study ii}(c), if we remove the jitter strategy of proposal boxes in PBR stage, the performance drops to 14.2 AP.

\begin{table}[tb!]
    \centering
\begin{minipage}[htb]{0.47\linewidth}
    \begin{minipage}[ht]{1.0\linewidth}
        \setlength{\tabcolsep}{3pt}
        \renewcommand\arraystretch{0.6}
        \centering
        \begin{tabular}{c|ccc}
        \specialrule{0.13em}{0pt}{1pt}
        Methods & AR$_{1}$ & AR$_{10}$ & AR$_{100}$  \\
        \specialrule{0.08em}{0pt}{0pt}
        UFO$^2$ & 14.7 & 22.6 & 23.3 \\
        P2BNet-FR & \textbf{21.3}  & \textbf{32.8}  &\textbf{34.2}\\
        \specialrule{0.13em}{0pt}{0pt}
        \end{tabular}
        %\vspace{-0.5em}
        \flushleft{\par{(a) Comparisons of average recall for UFO$^2$ and P2BNet-FR.}}
        % \label{tab:AR}
    \end{minipage}

    \begin{minipage}[h]{0.52\linewidth}
    \centering
        \begin{tabular}{c|cc}
        \specialrule{0.13em}{0pt}{0pt}
        Balance & AP & AP$_{50}$ \\
        \specialrule{0.08em}{0pt}{0pt}
        \checkmark     & \textbf{21.7}& \textbf{46.1}  \\
        -  & 12.9  & 36.0  \\
        \specialrule{0.13em}{0pt}{0pt}
        \end{tabular}
        % \caption{Caption}
        % \label{tab:unbalance analyse}
    %\vspace{-0.5em}
    \flushleft{\par{(b) Unbalance issue.}}
    % \label{tab:AR}
\end{minipage}
    \begin{minipage}[h]{0.46\linewidth}
        \centering
            \begin{tabular}{c|cc}
            \specialrule{0.13em}{0pt}{0pt}
            Jitter &AP & AP$_{50}$ \\
            \specialrule{0.08em}{0pt}{0pt}
            \checkmark   & \textbf{21.7}&\textbf{46.1}  \\
            -  &14.2 & 38.2 \\
            \specialrule{0.13em}{0pt}{0pt}
            \end{tabular}
        % \caption{Different backbone of P2BNet}
        %\vspace{-0.5em}
        \flushleft{\par{(c) Jitter strategy.}}
        % \label{tab:backbone}
\end{minipage}
\end{minipage}
%\vspace{0.5em}
\begin{minipage}[htb]{0.52\linewidth}
    \centering
    \begin{tabular}{c|cc|cc}
    \specialrule{0.13em}{0pt}{0pt}
    \multirow{2}{*}{Detectors} &  \multicolumn{2}{c|}{GT box} & \multicolumn{2}{c}{Pseudo box} \\
    \cline{2-5}
    &AP & AP$_{50}$&AP & AP$_{50}$\\
    \specialrule{0.08em}{0pt}{0pt}
    RetinaNet~\cite{DBLP:retinanet_focalloss} &36.5&55.4 &21.0 &44.9 \\
    Reppoint~\cite{DBLP:reppoint} &37.0 & 56.7 &20.8 &45.1 \\
    Sparse R-CNN~\cite{DBLP:sparsercnn} &\textbf{37.9} &56.0 &21.1&43.3\\
    FR-FPN~\cite{FasterRCNN,DBLP:FPN} &37.4&\textbf{58.1}&\textbf{22.1} &\textbf{47.3} \\
    % SparseRCNN & &      \\
    \specialrule{0.13em}{0pt}{0pt}
    \end{tabular}
    %\vspace{-0.5em}
    \flushleft{\par{(d) Performance of different detectors on ground-truth box annotations and pseudo boxes generated by P2BNet. We use the top-4 for box merging.}}
    % \label{tab:detectors}
\end{minipage}
\caption{Ablation study (Part II).}
\label{tab: ablation study ii}
%\vspace{-2em}
%\vspace{-0.5cm}
\end{table}

\noindent\textbf{Number of Refinements in PBR.} Refining pseudo boxes is a vital part of P2BNet, and the cascade structure is used for iterative refinement to improve performance. Tab.~\ref{tab: ablation stdudy i}(c) shows the effect of the refining number in the PBR stage. One refinement brings a performance gain of 8.0 AP, up to a competitive 21.7 AP. The highest 21.9 AP is obtained with two refinements, and the performance is saturated.
%Without refinement, \ie~only the CBP stage, performance is merely 13.7 AP.
We choose one refinement as the default configuration.

\noindent\textbf{Box Merging Policy.} We use the top-$k$ score average weight as our merging policy. We find that the hyper-parameter $k$ is slightly sensitive and can be easily generalized to other datasets, as presented in Tab.~\ref{tab: ablation stdudy i}(b), and only the top-$1$ or top-$few$ proposal box plays a leading role in box merging. The best performance is 22.1 AP and 47.3 AP$_{50}$ when $k=4$. 
% $k$ is little insensitive and can be easily generalized to other datasets.
The mIoU$_{pred}$ between the pseudo box and ground-truth box is 57.5. In inference, if bag score $\mathbf{S}$
%(=\mathbf{S}^{cls} \odot \mathbf{S}^{ins}$)
is replaced by classification score $\mathbf{S}^{cls}$ for merging, the performance drops to 17.4 AP (vs 21.7 AP).

\noindent\textbf{Average Recall.}
% In Tab.~\ref{tab: ablation study ii}(a), we compare the mean average recall (AR) of \otsp-based PSOD method UFO$^2$ and P2BNet-FR to verify our motivation. 
In Tab.~\ref{tab: ablation study ii}(a), the AR in UFO$^2$ is 23.3, indicating a higher missing rate. Whereas the P2BNet-FR obtains 34.2 AR, far beyond that of the UFO$^2$. It shows our OTSP-free method is better at finding objects. 

\noindent\textbf{Unbalance Sampling Analysis.} To demonstrate the effect of unbalance sampling, we sample different numbers of proposal boxes for each object and keep them constant in every epoch during the training period. The performance drops in Tab.~\ref{tab: ablation study ii}(b) suggests the negative impact of unbalanced sampling.

\noindent\textbf{Different Detectors.}
We train different detectors~\cite{FasterRCNN,DBLP:FPN,DBLP:retinanet_focalloss,DBLP:reppoint,DBLP:sparsercnn} for the integrity experiments, all of which are conducted on R-50, as shown in Tab.~\ref{tab: ablation study ii}(d). Our framework exhibits competitive performance on other detectors. Box supervised performances are listed to demonstrate the upper bound of our framework. 
% With the pseudo box of P2BNet, we conduct experiments on FR-FPN~\cite{FasterRCNN,DBLP:FPN}, RetinaNet~\cite{DBLP:retinanet_focalloss}, Reppoint~\cite{DBLP:reppoint}, and Sparse R-CNN~\cite{DBLP:sparsercnn}. 
 
%\vspace{-0.5cm}
% %%%%%%%%%%%%%%%%%%%%%%%%%%%%%%%%%%%%%%%%%%%%%%%
\section{Conclusion}
%\vspace{-0.3cm}
In this paper, we give an in-depth analysis of shortcomings in \otsp-based PSOD frameworks, and further propose a novel \otsp-free network termed P2BNet to obtain inter-objects balanced and high-quality proposal bags. The coarse-to-fine strategy divides the prediction of pseudo boxes into CBP and PBR stages. In the CBP stage, fixed sampling is performed around the annotated points, and coarse pseudo boxes are predicted through instance-level MIL. The PBR stage performs adaptive sampling around the estimated boxes to finetune the predicted boxes in a cascaded fashion. As mentioned above, P2BNet takes full advantage of point information to generate high-quality proposal bags, which is more conducive to optimizing the detector (FR). Remarkably, the conceptually simple P2BNet-FR framework yields state-of-the-art performance with single point annotation.

\noindent\textbf{Acknowledgements}
%\vspace{-0.3cm}
This work was supported in part by the Youth Innovation Promotion Association CAS, the National Natural Science Foundation of China (NSFC) under Grant No. 61836012, 61771447 and 62006244, the Strategic Priority Research Program of the Chinese Academy of Sciences under Grant No.XDA27000000, and Young Elite Scientist Sponsorship Program of China Association for Science and Technology YESS20200140.

\bibliographystyle{splncs04}
\bibliography{egbib}

\begin{thebibliography}{10}
\providecommand{\url}[1]{\texttt{#1}}
\providecommand{\urlprefix}{URL }
\providecommand{\doi}[1]{https://doi.org/#1}

\bibitem{DBLP:MCG}
Arbel{\'{a}}ez, P.A., Pont{-}Tuset, J., \etal, J.T.B.: Multiscale combinatorial
  grouping. In: CVPR (2014)

\bibitem{DBLP:wsddn}
Bilen, H., Vedaldi, A.: Weakly supervised deep detection networks. In: CVPR
  (2016)

\bibitem{DBLP:series/lncs/Bottou12}
Bottou, L.: Stochastic gradient descent tricks. In: Neural Networks: Tricks of
  the Trade - Second Edition. Springer (2012)

\bibitem{DBLP:DETR}
Carion, N., Massa, F., \etal, G.S.: End-to-end object detection with
  transformers. In: ECCV (2020)

\bibitem{mmdetection}
Chen, K., Wang, J., Pang, J.e.: {MMDetection}: Open mmlab detection toolbox and
  benchmark. arXiv preprint arXiv:1906.07155  (2019)

\bibitem{DBLP:slv}
Chen, Z., Fu, Z., \etal, R.J.: {SLV:} spatial likelihood voting for weakly
  supervised object detection. In: CVPR (2020)

\bibitem{cheng2021pointly}
Cheng, B., Parkhi, O., Kirillov, A.: Pointly-supervised instance segmentation.
  CoRR  (2021)

\bibitem{DBLP:wccn}
Diba, A., Sharma, V., \etal, A.M.P.: Weakly supervised cascaded convolutional
  networks. In: CVPR (2017)

\bibitem{DBLP:roitransformer}
Ding, J., Xue, N., Long, Y., Xia, G., Lu, Q.: Learning roi transformer for
  oriented object detection in aerial images. In: CVPR (2019)

\bibitem{DBLP:VOC}
Everingham, M., Gool, L.V., \etal, C.K.I.W.: The pascal visual object classes
  {(VOC)} challenge. IJCV  (2010)

\bibitem{DBLP:C-WSL}
Gao, M., Li, A., \etal, R.Y.: {C-WSL:} count-guided weakly supervised
  localization. In: ECCV (2018)

\bibitem{MEFF}
Ge, W., Yang, S., Yu, Y.: Multi-evidence filtering and fusion for multi-label
  classification, object detection and semantic segmentation based on weakly
  supervised learning. In: CVPR (2018)

\bibitem{fastrcnn}
Girshick, R.B.: Fast {R-CNN}. In: ICCV (2015)

\bibitem{DBLP:beyond_bbox}
Guo, Z., Liu, C., Zhang, X., Jiao, J., Ji, X., Ye, Q.: Beyond bounding-box:
  Convex-hull feature adaptation for oriented and densely packed object
  detection. In: CVPR (2021)

\bibitem{DBLP:MASK-RCNN}
He, K., Gkioxari, G., \etal, P.D.: Mask {R-CNN}. In: ICCV (2017)

\bibitem{DBLP:resnet}
He, K., Zhang, X., \etal, S.R.: Deep residual learning for image recognition.
  In: CVPR (2016)

\bibitem{CASD}
Huang, Z., Zou, Y., \etal, B.V.K.V.K.: Comprehensive attention
  self-distillation for weakly-supervised object detection. In: NeurIPS (2020)

\bibitem{DBLP:guidingnet}
Jia, Q., Wei, S., \etal, T.R.: Gradingnet: Towards providing reliable
  supervisions for weakly supervised object detection by grading the box
  candidates. In: AAAI (2021)

\bibitem{DBLP:ANTI-UAV}
Jiang, N., Wang, K., Peng, X., Yu, X., Wang, Q., Xing, J., Li, G., Zhao, J.,
  Guo, G., Han, Z.: Anti-uav: {A} large multi-modal benchmark for {UAV}
  tracking. {IEEE} TMM  (2021)

\bibitem{DBLP:alexnet}
Krizhevsky, A., Sutskever, I., Hinton, G.E.: Imagenet classification with deep
  convolutional neural networks. In: NIPS (2012)

\bibitem{Lee_2021_ICCV}
Lee, P., Byun, H.: Learning action completeness from points for
  weakly-supervised temporal action localization. In: ICCV (2021)

\bibitem{DBLP:FPN}
Lin, T., Doll{\'{a}}r, P., \etal, R.B.G.: Feature pyramid networks for object
  detection. In: CVPR (2017)

\bibitem{DBLP:retinanet_focalloss}
Lin, T., Goyal, P., \etal, R.B.G.: Focal loss for dense object detection. In:
  ICCV (2017)

\bibitem{coco}
Lin, T.Y., Maire, M.e.: Microsoft coco: Common objects in context. In: ECCV
  (2014)

\bibitem{DBLP:SSD}
Liu, W., Anguelov, D., \etal, D.E.: {SSD:} single shot multibox detector. In:
  ECCV (2016)

\bibitem{DBLP:swin-transformer}
Liu, Z., Lin, Y., \etal, Y.C.: Swin transformer: Hierarchical vision
  transformer using shifted windows. In: ICCV (2021)

\bibitem{DBLP:journals/tip/MengZYZZW22}
Meng, M., Zhang, T., Yang, W., Zhao, J., Zhang, Y., Wu, F.: Diverse
  complementary part mining for weakly supervised object localization. {IEEE}
  TIP  (2022)

\bibitem{Click}
Papadopoulos, D.P., Uijlings, J.R.R., \etal, F.K.: Training object class
  detectors with click supervision. In: CVPR (2017)

\bibitem{DBLP:yolo}
Redmon, J., Divvala, S.K., \etal, R.B.G.: You only look once: Unified,
  real-time object detection. In: CVPR (2016)

\bibitem{FasterRCNN}
Ren, S., He, K., \etal, R.B.G.: Faster {R-CNN:} towards real-time object
  detection with region proposal networks. {IEEE} TPAMI  (2017)

\bibitem{ICMWSD}
Ren, Z., Yu, Z., \etal, X.Y.: Instance-aware, context-focused, and
  memory-efficient weakly supervised object detection. In: CVPR (2020)

\bibitem{UFO2}
Ren, Z., Yu, Z., \etal, X.Y.: Ufo\({}^{\mbox{2}}\): {A} unified framework
  towards omni-supervised object detection. In: ECCV (2020)

\bibitem{DBLP:withoutboundingbox}
Ribera, J., Guera, D., Chen, Y., Delp, E.J.: Locating objects without bounding
  boxes. In: CVPR (2019)

\bibitem{DBLP:selectivesearch}
van~de Sande, K.E.A., Uijlings, J.R.R., \etal, T.G.: Segmentation as selective
  search for object recognition. In: ICCV (2011)

\bibitem{DBLP:uwsod}
Shen, Y., Ji, R., Chen, Z., Wu, Y., Huang, F.: {UWSOD:} toward
  fully-supervised-level capacity weakly supervised object detection. In:
  NeurIPS (2020)

\bibitem{DBLP:VGG}
Simonyan, K., Zisserman, A.: Very deep convolutional networks for large-scale
  image recognition. In: ICLR (2015)

\bibitem{p2pnet}
Song, Q., Wang, C., Jiang, Z., Wang, Y., Tai, Y., Wang, C., Li, J., Huang, F.,
  Wu, Y.: Rethinking counting and localization in crowds: A purely point-based
  framework. In: ICCV (2021)

\bibitem{DBLP:sparsercnn}
Sun, P., Zhang, R., \etal, Y.J.: Sparse {R-CNN:} end-to-end object detection
  with learnable proposals. In: CVPR (2021)

\bibitem{DBLP:oicr}
Tang, P., \etal, X.W.: Multiple instance detection network with online instance
  classifier refinement. In: CVPR (2017)

\bibitem{DBLP:pcl}
Tang, P., Wang, X., \etal, S.B.: {PCL:} proposal cluster learning for weakly
  supervised object detection. {IEEE} TPAMI  (2020)

\bibitem{DBLP:MELM}
Wan, F., Wei, P., \etal, Z.H.: Min-entropy latent model for weakly supervised
  object detection. {IEEE} TPAMI  (2019)

\bibitem{C-MIDN}
Yan, G., Liu, B., \etal, N.G.: {C-MIDN:} coupled multiple instance detection
  network with segmentation guidance for weakly supervised object detection.
  In: ICCV (2019)

\bibitem{DBLP:r3det}
Yang, X., Yan, J., Feng, Z., He, T.: R3det: Refined single-stage detector with
  feature refinement for rotating object. In: AAAI (2021)

\bibitem{DBLP:reppoint}
Yang, Z., Liu, S., \etal, H.H.: Reppoints: Point set representation for object
  detection. In: ICCV (2019)

\bibitem{CPR}
Yu, X., Chen, P., \etal, D.W.: Object localization under single coarse point
  supervision. In: CVPR (2022)

\bibitem{scalematch}
Yu, X., Gong, Y., \etal, N.J.: Scale match for tiny person detection. In: IEEE
  WACV (2020)

\bibitem{WSOD2}
Zeng, Z., Liu, B., \etal, J.F.: {WSOD2:} learning bottom-up and top-down
  objectness distillation for weakly-supervised object detection. In: ICCV
  (2019)

\bibitem{DBLP:weaklysurvey}
Zhang, D., Han, J., Cheng, G., Yang, M.: Weakly supervised object localization
  and detection: {A} survey. {IEEE} TPAMI  (2021)

\bibitem{DBLP:Acol}
Zhang, X., Wei, Y., \etal, J.F.: Adversarial complementary learning for weakly
  supervised object localization. In: CVPR (2018)

\bibitem{DBLP:2ND_ANTI-UAV}
Zhao, J., Wang, G., Li, J., Jin, L., Fan, N., Wang, M., Wang, X., Yong, T.,
  Deng, Y., Guo, Y., Ge, S., Guo, G.: The 2nd anti-uav workshop {\&} challenge:
  Methods and results. ICCVW 2021  (2021)

\bibitem{DBLP:cam}
Zhou, B., Khosla, A., \etal, {\`{A}}.L.: Learning deep features for
  discriminative localization. In: CVPR (2016)

\bibitem{DBLP:DeformableDETR}
Zhu, X., Su, W., \etal, L.L.: Deformable {DETR:} deformable transformers for
  end-to-end object detection. In: ICLR (2021)

\bibitem{DBLP:EdgeBox}
Zitnick, C.L., Doll{\'{a}}r, P.: Edge boxes: Locating object proposals from
  edges. In: ECCV (2014)

\end{thebibliography}

\clearpage

\begin{center}{\bf \Large Appendix}\end{center}%\vspace{-2mm}
\section{Implementation Details for P2BNet}
\subsection{Sampling Settings.} 
In CBP sampling, $s\in \{4, 8, 16, 32, 64, 128\} \cdot \delta$, where $\delta$ is a factor for dynamic adjustment according to the dataset. $\delta=min(W,H)/100$, where $W$ and $H$ are width and height of the image. $v\in \{1/3, 1/2, 2/3, 1, 3/2, 2, 3\}$ is used as the fixed setting. In PBR sampling, we deem $(v\cdot s)$ and $(v/s)$ as a whole respectively, and set $(v\cdot s) \in \{0.7, 0.8, 1, 1.2, 1.3\}$, $(v/s) \in \{0.7, 0.8, 1, 1.2, 1.3\}$, so we have $5\times5=25$ options. $(o_x,o_y) \in \{(0,0), (1,0), (0,1), (-1,0), (-1,-1)\}$ is used to jitter the center position. These settings are simple and fixed, which is beneficial for better generation. In negative sampling, we randomly sample 500 boxes, filter out those which have high IoU with all positive proposals and obtain the final negative sample set $\mathcal{N}$.

\subsection{Other Experimental Settings.} 
ResNet-50 is used as the backbone network unless otherwise specified. The FPN structure is also utilized. The mini-batch is 16 images and all models are trained with 8 GPUs for the COCO dataset. In Tab.~\ref{tab: ablation stdudy i} (a,c), the number of PBR stages has been described in the corresponding section. Loss weights are set as $\alpha_{mil1}=0.25$ in CBP, $\alpha_{mil2}=0.25$ and $\alpha_{neg}=0.75$ in PBR, which follow focal loss~\cite{DBLP:retinanet_focalloss} and are fixed during training. Unless otherwise specified, we use one CBP stage and one PBR stage as our default configuration, which the experiments of Tab.~\ref{tab: ablation stdudy i} (b) and~\ref{tab: ablation study ii} (a,b,c) depend on. 

\section{Other Experiments.} 
% Some additional experiments are supplemented here in this section.

\subsection{Different Backbones of P2BNet.}
We change the backbone to a larger network, ResNet-101, but the performance decreases. The performance of R-50 is 21.7 AP and 46.1 AP$_{50}$ while that of R-101 is 20.8 AP and 45.8 AP$_{50}$. We conjecture this is because the larger the network, the more possible it is to predict the discriminative part rather than the whole object. This phenomenon also happens in WSOD. 

\subsection{Annotation Time of Quasi-center Point.}
Quasi-center point annotation requires annotating objects in center region. Generally, center region of the object is the body part which has more saliency and easy to annotate. For comparison, 200 images were manually annotated by three annotators. The average annotation time per object is 4.27s for quasi-center strategy and 3.84s for the previous random annotation style. Considering the huge performance gain, we believe this is a fair trade-off. 

% In Tab.~\ref{tab: ablation study ii}(---), the AP$_{50}$ is similar in R-50 and R-101 backbone, while the AP$_{75}$ of R-101 is lower.
% % , showing that the larger network is more likely to predict the discriminative part.

\section{The visualization of P2BNet-FR}
\subsection{Visualization of P2BNet.}
\begin{figure}
    \centering
    \includegraphics[width=1.\linewidth]{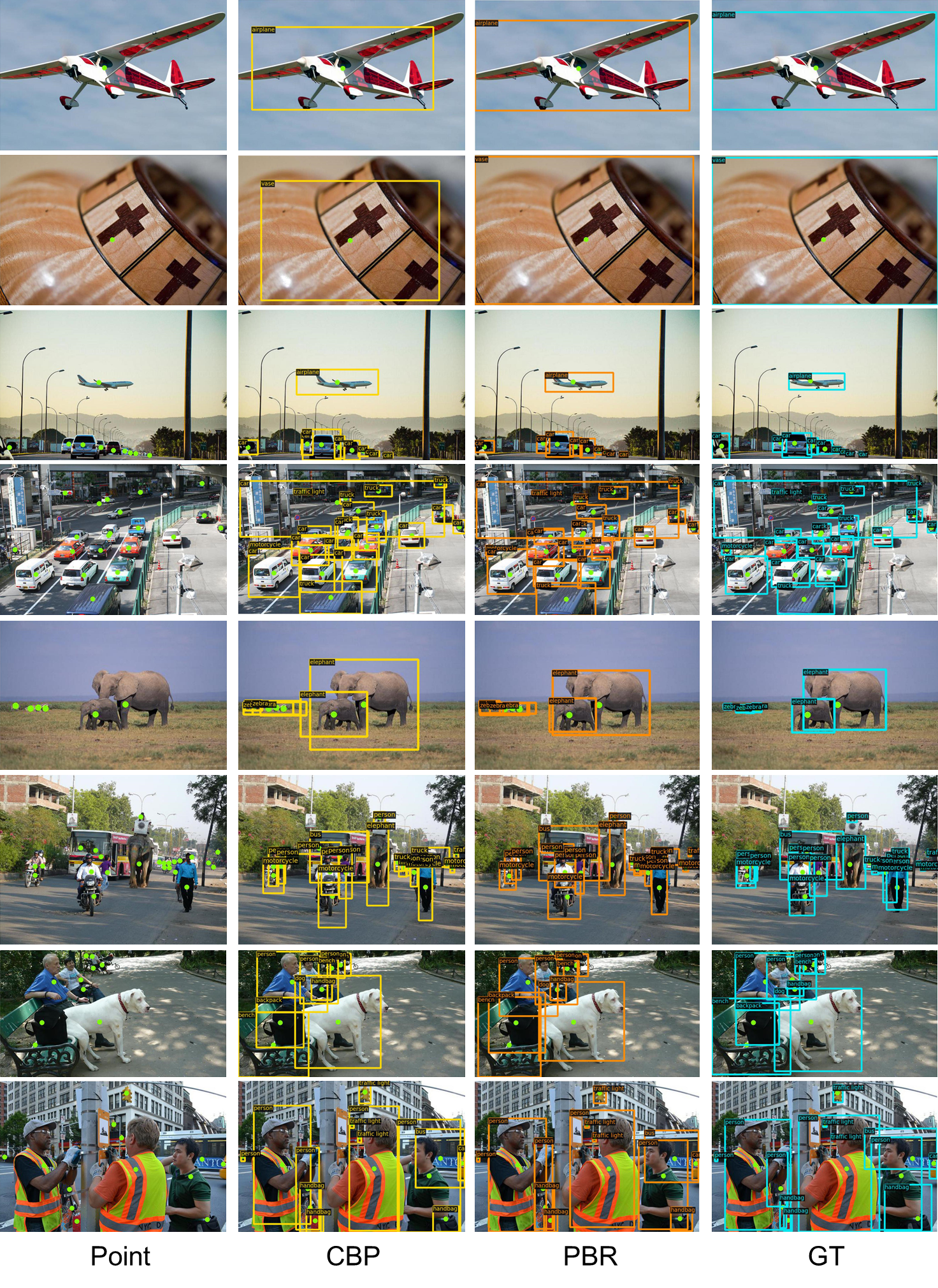}
    \caption{The Visualization of P2BNet. The green, yellow, orange and blue represent the annotation point, the CBP result, the PBR result and the ground-truth, respectively. With CBP stage and PBR stage, we obtain the high-quality pseudo boxes. The performance in dense scenarios is also great. The images are from COCO-17 train set. (Best viewed in color)}
    \label{fig:vis_p2b}
\end{figure}

In Fig.~\ref{fig:vis_p2b}, we illustrate the visualization of P2BNet. With QC point annotation, P2BNet predicts coarse pseudo boxes in the CBP stage, and refines the quality of estimated boxes in the PBR stage. 
% We find the IoU of results with the ground-truth in the PBR stage is better. 
Furthermore, Fig.~\ref{fig:vis_p2b} also shows the performance of P2BNet in complex scenarios, reflecting that the pseudo boxes predicted by P2BNet have the ability to well represent their respective targets.  
% Our P2BNet can predict great pseudo boxes in the images with dense objects to represent respective targets well. 
The estimated boxes are used as the pseudo annotation to train the detector, so the high quality of pseudo boxes guarantees high detection performance.

\subsection{Visualization of Detection Performance.}
\begin{figure}
    \centering
    \includegraphics[width=1.\linewidth]{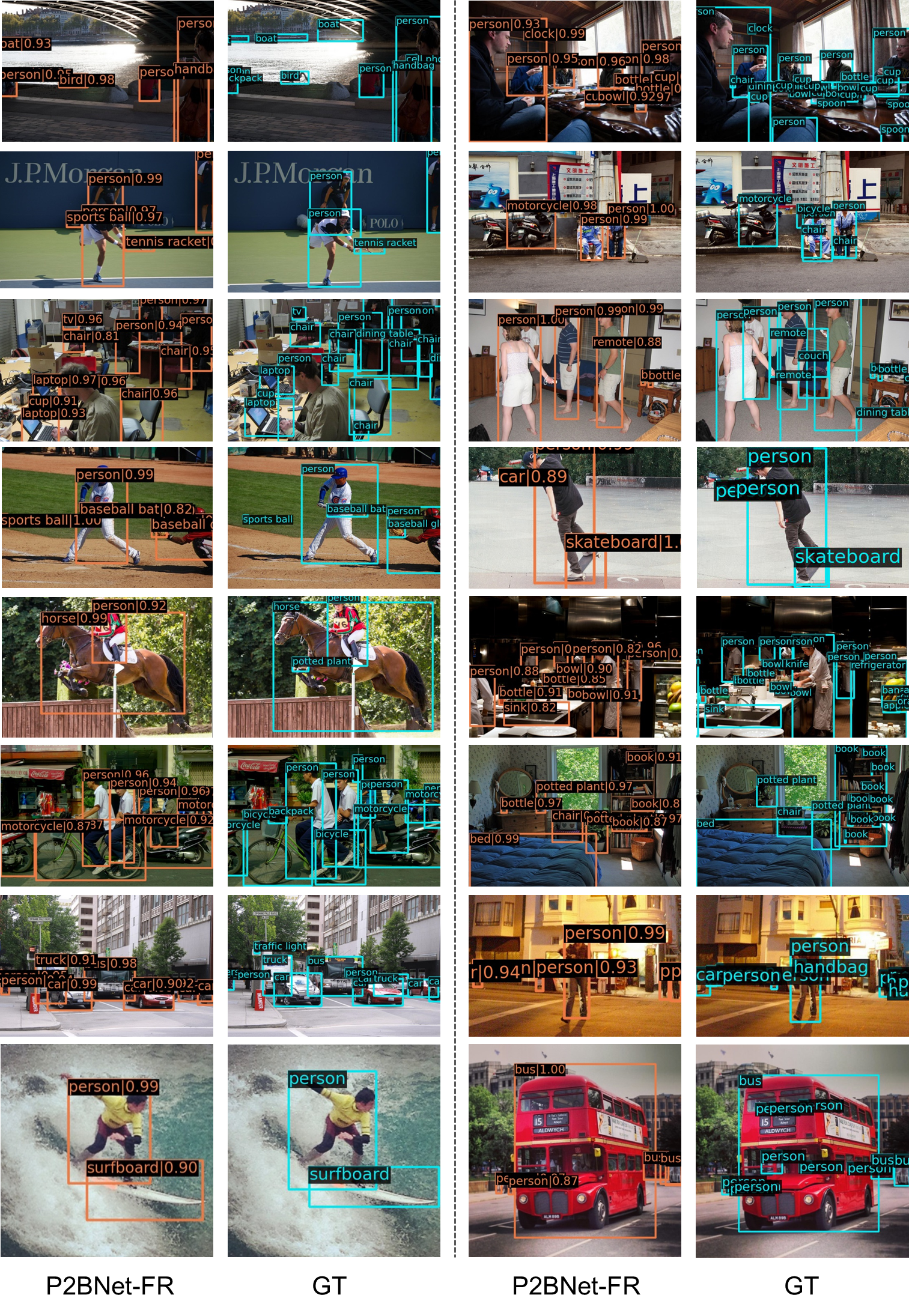}
    \caption{The detection visualization of P2BNet-FR. Orange and blue represent detection result and ground-truth. Our P2BNet-FR can detect objects in complex scenarios with point annotation. The images are from COCO-17 val set. (Best viewed in color)}
    \label{fig:vis_p2bfr}
\end{figure}

Some other detection results of P2BNet-FR are given in Fig.~\ref{fig:vis_p2bfr}. It shows our PSOD detector's performance is comparable with that under box supervision. The performance of WSOD and PSOD is low in complex dataset like COCO. However, our PSOD detector P2BNet-FR has achieved great improvement, especially in complex and dense scenarios. 
\end{document}